\documentclass{article}
\PassOptionsToPackage{numbers}{natbib}

 \usepackage[eandd, final]{neurips_2026}

\usepackage[utf8]{inputenc} 
\usepackage[T1]{fontenc}    
\usepackage{hyperref}       
\usepackage{url}            
\usepackage{booktabs}       
\usepackage{amsfonts}       
\usepackage{nicefrac}       
\usepackage{microtype}      
\usepackage{xcolor}         
\usepackage{xspace}
\usepackage{caption}
\usepackage{graphicx}
\usepackage{todonotes}
\usepackage{wrapfig}
\usepackage{subcaption}
\usepackage{enumitem}
\usepackage{fvextra}
\usepackage{upquote}
\usepackage{fontawesome5}
\usepackage{booktabs}
\usepackage{pifont}
\newcommand{\cmark}{\ding{51}}
\newcommand{\xmark}{\ding{55}}

\newcommand{\DatasetName}{\textsc{$\mathbb{M}$ARBLE}\xspace}
\newcommand{\Mcube}{\textsc{Cube}\xspace}

\newcommand{\Mmaze}{\textsc{Maze}\xspace}
\newcommand{\Mbloxorz}{\textsc{Blox}\xspace}

\hypersetup{hidelinks}

\definecolor{TblHead}{HTML}{E9EEF7}   
\definecolor{TblStripe}{HTML}{F6F8FB} 

\definecolor{lightgray}{gray}{0.9}
\definecolor{OpenSrcHead}{gray}{0.8}
\definecolor{ClosedSrcHead}{gray}{0.8}
\definecolor{stratIgnored}{HTML}{999999}   
\definecolor{stratHarness}{HTML}{636E72}   
\definecolor{stratSingle}{HTML}{E63946}    
\definecolor{stratDistrust}{HTML}{D48B2F}  
\definecolor{stratIterative}{HTML}{2A9D8F} 
\definecolor{errInvalid}{HTML}{7096DD}
\definecolor{errMechanics}{HTML}{54B7AC}
\definecolor{errFeedback}{HTML}{F1B66C}
\definecolor{errSearch}{HTML}{A58BCD}

\usepackage[most]{tcolorbox}

\newtcolorbox{findingbox}[1][]{
  enhanced,
  colback=white,
  colframe=teal!70!black,
  arc=0.5mm,
  boxrule=1pt,
  drop shadow={shadow xshift=1.5mm, shadow yshift=-1.5mm, opacity=0.3},
  left=10pt,
  right=10pt,
  top=6pt,
  bottom=6pt,
  before upper={\raisebox{-0.6em}{\includegraphics[height=1.8em]{figures/icons/marble_icon.png}}}
}

\global
\tcbset{
  aibox/.style={
    width=\textwidth,
    top=10pt,
    colback=white,
    colframe=black,
    colbacktitle=black,
    center,
  }
}

\tcbset{
  emptyBox/.style={
    width=\textwidth,
    top=-0.7cm,
    colback=white,   
    colframe=white,  
    boxrule=0pt,     
    center,
    notitle,         
  }
}
\newtcolorbox{AIboxICLR}[2][]{emptyBox,title=#2,#1}

\newtcolorbox{AIbox}[2][]{aibox,title=#2,#1}
\tcbset{
  aiboxsmall/.style={
    width=0.62\textwidth,
    top=10pt,
    colback=white,
    colframe=black,
    colbacktitle=black,
    enhanced,
    top,
    attach boxed title to top left={yshift=-0.1in,xshift=0.15in},
    boxed title style={boxrule=0pt,colframe=white,},
  }
}
\newtcolorbox{AIboxSmall}[2][]{aiboxsmall,title=#2,#1}

\definecolor{OpenSrcHead}{HTML}{CFE6FF}
\definecolor{OpenSrcStripe}{HTML}{E2EEFF}   
\definecolor{ClosedSrcHead}{HTML}{FFDDBD}
\definecolor{ClosedSrcStripe}{HTML}{FFE8D2} 
\definecolor{OpenSrcRow}{HTML}{F3F9FE}    
\definecolor{ClosedSrcRow}{HTML}{FFF4E9}  

\usepackage{array}

\usepackage[most]{tcolorbox} 

\newtcolorbox{promptbox}[1]{
    colback=white,
    colframe=black,
    colbacktitle=black,
    coltitle=white,
    fonttitle=\bfseries,
    title=#1,
    boxrule=0.8pt,
    arc=2pt,
    left=10pt,
    right=10pt,
    top=8pt,
    bottom=8pt,
    before skip=8pt,
    after skip=12pt
}

\title{A Benchmark of Agentic Spatial Intelligence for Multimodal LLMs}
\title{MARBLE: An Agentic Benchmark for Spatial Reasoning and Visual Abstraction}

\definecolor{linkblue}{HTML}{2878D0}

\author{
\makebox[\dimexpr\textwidth-2\tabcolsep\relax][c]{%
\normalfont
\begin{tabular}[t]{@{}c@{}}
\textbf{Yulun Jiang}$^{1,2}$ \quad
\textbf{Naël Ouerghemi}$^{1}$ \quad
\textbf{Yekun Chai}$^{2}$ \quad
\textbf{Maria Brbi\'{c}}$^{1}$ \quad
\textbf{Michael Moor}$^{2}$ \\[6pt]
$^{1}$EPFL \quad $^{2}$ETH Zurich \\[6pt]
\href{https://marble-benchmark.github.io/}{
\textcolor{linkblue}
{\texttt{
    \faGlobe\enspace marble-benchmark.github.io
    }}
}
\end{tabular}
}
}

\begin{document}

\maketitle

\begin{abstract}

A central challenge for multimodal language models (MLLMs) in real-world tasks is to abstract visual observations into structured spatial states and reason over how those states change under actions over long horizons. 
Existing multimodal benchmarks often focus on static question answering, where answers can be directly extracted from images, or assume symbolic states are available and bypass the visual-to-spatial abstraction process.
We present \DatasetName, a diagnostic benchmark for spatially grounded multimodal reasoning and planning across three agent environments: \Mcube, \Mmaze, and \Mbloxorz.
\DatasetName spans 2D and 3D spatial puzzles under geometric constraints, dynamic layouts, and spatial transformations, with difficulty systematically controlled along two axes: observation format and planning horizon.
Extensive evaluations show that frontier MLLMs struggle on \DatasetName, with the strongest model GPT-5.4 achieving only a 7.3\% average success rate across the three environments in the hardest setting.
Pairwise ablations on the open-weight models identify visual-to-spatial abstraction as the major bottleneck: when models must infer spatial states from images instead of receiving human-annotated states, success drops from 50.7\% to 3.7\% for Gemma-4-31B and from 87.9\% to 6.5\% for Kimi-K2.6.
The results indicate that models often fail to construct precise spatial states from visual observations before planning, a task that is straightforward for humans.
Overall, \DatasetName reveals the gap in reliable visual-to-spatial abstraction for frontier MLLMs and provides a controlled testbed for measuring progress in spatially grounded multimodal reasoning and planning.

\end{abstract}
\begin{figure}[h]
    \includegraphics[width=\linewidth]{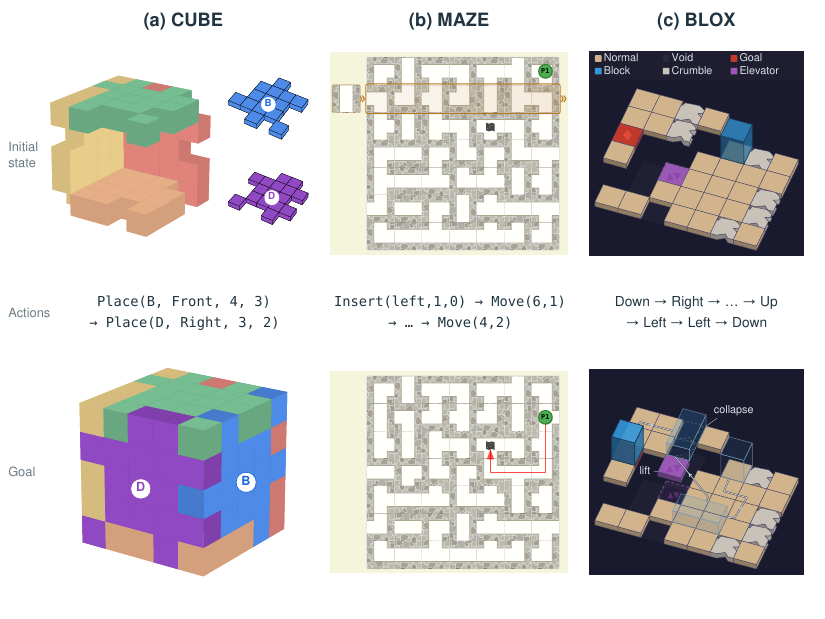}
    \vspace{-1em}
    \caption{
    \DatasetName{} contains three spatially grounded environments: \Mcube{}, \Mmaze{}, and \Mbloxorz{}, designed to evaluate MLLMs' multimodal reasoning and planning abilities under geometric constraints, dynamic connectivity, and action-conditioned spatial transformations.
    }
    \vspace{-1.5em}
    \label{fig:teaser}
\end{figure}

\section{Introduction}
\label{sec:introduction}

Multimodal large language models (MLLMs) are expected to perceive, reason, and act in environments where visual observations ground long-horizon decision-making.
In spatially grounded tasks, this requires visual-to-spatial abstraction: converting visual observations into structured spatial states, tracking how those states transform under actions, and leveraging them for planning. 
However, current MLLMs are still primarily trained and evaluated on language-centric tasks, leaving their ability to perform spatial world modeling in agentic tasks poorly understood~\cite{liu2023visual,alayrac2022flamingo,tong2024cambrian,yang2025cambrian}. Developing this capability is critical for building intelligent systems that can generalize across domains and adaptively interact with complex environments~\cite{ha2018world,wang2025vagen,cheng2025spa}.

Recent benchmarks have shown that MLLMs can solve tasks involving both visual understanding and reasoning in various domains~\cite{lu2022learn,lu2023mathvista, yue2024mmmu}. 
However, these benchmarks often focus on visual question answering in static scenes, where the answer can often be directly extracted from the visual input.
There have been other benchmarks evaluating MLLMs' spatial reasoning abilities~\cite{kamath2023s,liu2023visual,yang2024vsibench,li2025stare}, but they are typically single-turn and do not require models to update a spatial state across action-conditioned transitions. 
More recent game-inspired agentic benchmarks have begun to evaluate MLLMs' capability in solving multi-turn interactive tasks~\cite{wu2024vsp,wang2025multimodalescape,wang2026visgym,li2026spatialworldmodels}.
However, they either operate over simplified symbolic or grid representations, or emphasize task success without diagnostic controls that isolate visual-to-spatial abstraction from long-horizon transition reasoning. As a result, it remains unclear whether the current MLLMs can construct useful spatial states from visual input and use them to reason and plan in spatially grounded environments.

In this work, we present \DatasetName{} (MultimodAl Reasoning Benchmark for Language modEls), a diagnostic benchmark for visually grounded spatial reasoning and planning across three environments: \Mcube{}, \Mmaze{}, and \Mbloxorz{}. Each environment requires agents to infer a task-relevant spatial state from visual observations and then plan through action-induced transformations. The three environments cover complementary challenges: \Mcube{} emphasizes 3D orientation and geometric assembly, \Mmaze{} tests layout abstraction and planning under dynamic connectivity, and \Mbloxorz{} captures pose tracking and action-conditioned transitions under geometric constraints.

A key goal of \DatasetName{} is to provide a diagnostic evaluation. Rather than measuring only final task success, \DatasetName{} systematically controls difficulty along two axes: observation format and planning horizon. 
The observation format varies between visual-only observations and symbolic-augmented observations, enabling us to disentangle failures in visual abstraction from failures in reasoning and planning.
The planning horizon is defined by the length of the shortest solution, allowing us to assess how performance changes as models reason over longer sequences of action-conditioned state transitions.
Together, these controls move beyond measuring whether models solve a task and help identify where failures arise within the agentic reasoning pipeline.

We conduct extensive evaluations of ten advanced MLLMs on \DatasetName. The results show that frontier models can solve most tasks in the easy setting, with GPT-5.4 achieving the highest average success rate of 94.6\% across the three environments.
In contrast, all models struggle substantially in the hard setting, where GPT-5.4 achieves an average success rate of only 7.3\% across the three environments, and every model achieves 0.0\% success on \Mcube.
These results show that complex spatial reasoning and planning remain challenging for frontier MLLMs. 
To identify the source of these failures, we further conduct controlled experiments that disentangle the effects of planning horizon and visual-to-spatial abstraction.
We find that increasing the planning horizon mainly affects weaker models like Gemma-4-31B, causing a performance drop from 50.7\% to 28.5\% on average, while stronger models such as Kimi-K2.6 remain largely unaffected at around 88.0\% success rate.
In comparison, removing the symbolic states induces a much larger and more consistent degradation across models, reducing the performance from 50.7\% to 3.7\% for Gemma-4-31B and from 87.9\% to 6.5\% for Kimi-K2.6.
These findings indicate that visual-to-spatial abstraction is the major bottleneck: even frontier MLLMs often fail to construct precise spatial states from visual inputs, despite the same task being very simple for humans.
Overall, \DatasetName serves as a diagnostic testbed for systematically evaluating whether MLLMs can abstract visual observations into spatial states, update those states through action-conditioned transitions, and use them for long-horizon planning.

\section{Related Work}
\label{sec:related_work}

\paragraph{Multimodal Reasoning and Agentic MLLMs.}
Recent MLLMs have shown rapid progress on multimodal understanding and reasoning~\cite{hurst2024gpt, liu2023visual, li2024llava, bai2025qwen3,tong2024cambrian}, extending language-model reasoning techniques such as chain-of-thought, tool use, and test-time scaling to visual inputs~\cite{wei2022chain,kojima2022large,guo2025deepseek,wang2025multimodal,schick2023toolformer,yao2022react}.
Beyond static image question answering, current systems increasingly target agentic tasks where models interact with environments, receive feedback, and revise their actions over multiple turns~\cite{zheng2024gpt,hong2024cogagent,team2026kimi};
Vision-Language-Action models (VLAs) further extend this progress to embodied action spaces~\cite{black2024pi_0,bjorck2025gr00t,kim2024openvla,niu2024llarva,driess2023palm,zitkovich2023rt}.
However, these advances do not by themselves ensure robust visually grounded reasoning: current models remain largely language-centric, often reducing visual inputs to local attributes, captions, or salient cues rather than structured, action-relevant world states~\cite{tong2024eyes,fu2024blink,rahmanzadehgervi2024vision, asadi2026mirage}.
In contrast, \DatasetName{} evaluates whether multimodal agents can transform visual observations into spatial states and use those states to plan through long action sequences.

\paragraph{Spatial Reasoning and Visual Abstraction.}
Spatial reasoning is a long-standing problem in cognitive science and has become a central capability for evaluating multimodal intelligence~\cite{shepard1971mental,waller2013handbook}.
Recent studies show that MLLMs exhibit emerging visual-spatial abilities, including reasoning about spatial relations, viewpoints, scene layouts, and mental maps from visual observations~\cite{yang2024vsibench,cai2025easi,stogiannidis2025mindthegap,xu2025defining}.
A common theme across these works is that spatial reasoning depends on an intermediate representation: models must abstract raw visual inputs into structured spatial states, cognitive maps, or spatial mental models that support consistent reasoning beyond local perception~\cite{yin2025spatialmentalmodelinglimited,li2026spatialworldmodels}.
In agentic spatial tasks, this abstraction step becomes the basis for planning, because the agent must reason not only about the current spatial state but also about how actions transform that state over time~\cite{wang2025vagen,cheng2025spa}.
\DatasetName{} targets this bottleneck directly by comparing raw visual observations with symbolic spatial-state annotations.

\paragraph{Benchmarks for Multimodal and Spatial Reasoning.}
Existing multimodal benchmarks evaluate broad visual-language understanding and reasoning across diverse tasks and domains~\cite{yue2024mmmu,wang2024mmlu,lu2023mathvista,hao2025can,chen2024mega,chia2024puzzlevqa,chollet2024arc}.
Spatial reasoning benchmarks further probe models' ability to understand visual relations, geometric structure, perspective, and spatial mental models~\cite{kamath2023s,fu2024blink,yang2024vsibench,li2025stare,cai2025easi}.
However, these benchmarks primarily focus on visual question answering or short-form spatial reasoning, where models reason over a fixed observation rather than maintaining and updating an environment state through action-conditioned transitions.
Recent agentic and game-inspired benchmarks introduce interaction, feedback, or multi-step decision-making in visual environments~\cite{wang2025multimodalescape,paglieri2024balrog,zheng2025v,liu2024visualagentbench}.
Concurrent work, VisGym~\cite{wang2026visgym}, provides a suite of customizable environments for training and evaluating multimodal agents.
While these benchmarks advance multimodal agent evaluation, they serve as broad testbeds rather than focused diagnostics of spatial abstraction and action-conditioned reasoning.
\DatasetName{} fills this gap by a factorized analysis that systematically controls the observation format and reasoning depth in long-horizon spatial planning tasks.

\section{\DatasetName Benchmark}

\begin{figure}[h]
    \centering
     \includegraphics[width=0.95\linewidth]{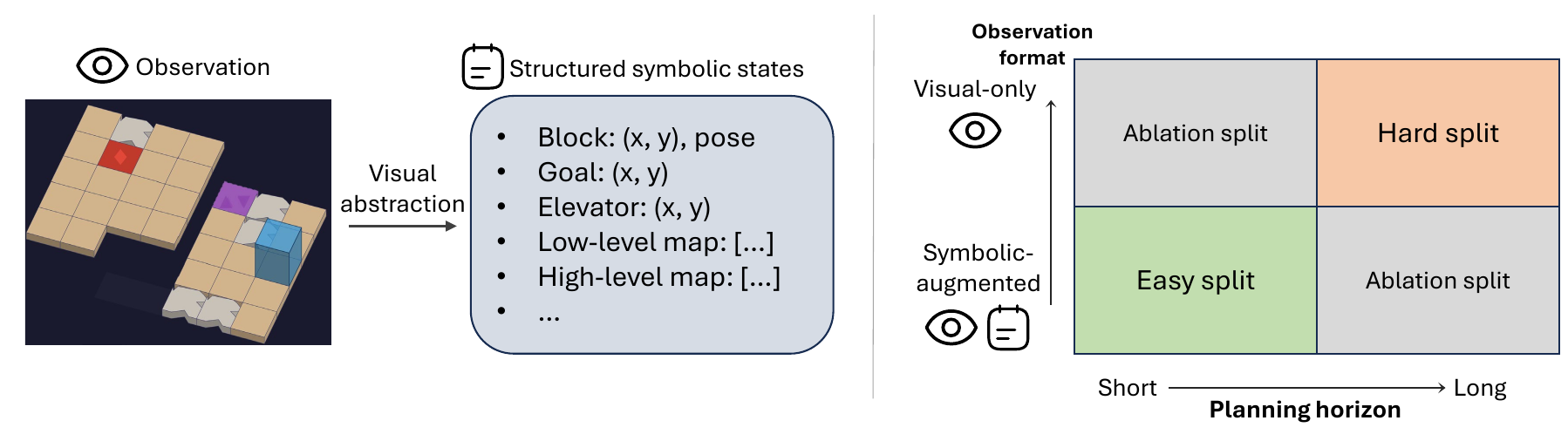}
    \caption{
    \textit{Left}: \textbf{Visual Abstraction} converts rendered observations into structured spatial states for downstream reasoning and planning.
    \textit{Right}: \DatasetName{} controls difficulty along two axes, observation format and planning horizon, yielding easy/hard benchmark splits and matched probes that isolate visual abstraction from long-horizon planning. Figure~\ref{fig:task_illustration} in Appendix~\ref{app:task_example} illustrates easy and hard task instances and their solution trajectories for each environment.
    }
    \label{fig:difficulty}
    \vspace{-3.0em}
\end{figure}

\textbf{Overview.}
\DatasetName contains three spatially grounded environments: \Mcube, \Mmaze, and \Mbloxorz.
Each task requires an agent to solve a multi-step reasoning and planning problem under explicit spatial constraints.
In \Mcube, the agent assembles a valid 3D cube from jigsaw-like face pieces by matching edge geometry and orientation; in \Mmaze, it plans through a shifting 2D labyrinth in which tile insertions alter board connectivity; and in \Mbloxorz, it navigates a block across a 3D-like board while tracking both its position and orientation, as illustrated in Figure~\ref{fig:teaser}.
Together, \DatasetName provides three gym-style environments covering geometric assembly, dynamic topology, and spatial transformation, which provides a controlled testbed for visual abstraction and planning in spatially grounded settings.
Detailed explanation of each environment is provided in Appendix~\ref{sup:environments}.

\textbf{State Observation and Visual Abstraction.}
At each turn, the agent receives a rendered observation of the current environment state together with task instructions.
To solve the task, the agent must extract task-relevant spatial structure from the observation, such as piece orientations, maze connectivity, or block pose.
We refer to this process as a \textit{visual abstraction}: converting raw visual input into a structured spatial state that can support downstream planning.
Visual abstraction is a core entry point for spatial world modeling, and prior work has shown that multimodal models often struggle to separate spatial perception from reasoning~\cite{fu2024blink}.
To isolate the role of visual abstraction, \DatasetName provides an optional human-readable symbolic representation alongside the rendered observation.
When this symbolic representation is provided, the agent can rely on an explicit structured description of the current state instead of having to infer all relevant spatial structure from visual input.
Comparing visual-only and symbolic-augmented observations therefore reveals how much model performance depends on successful visual abstraction.

\textbf{Difficulty Control.}
\DatasetName{} controls difficulty along two diagnostic axes: \textit{observation format} and \textit{planning horizon}.
The observation format varies between visual-only observations and symbolic-augmented observations.
In the visual-only setting, the agent must infer the task-relevant spatial state from the rendered input; in the symbolic-augmented setting, the rendered input is supplemented with a human-readable symbolic state description (Figure~\ref{fig:difficulty} left).
The planning horizon is defined as the length of the shortest solution, capturing the number of action-conditioned state transitions the agent must reason over.
As the horizon increases, the number of candidate action sequences grows combinatorially, making the task substantially harder even when the relevant state is available.

For the main benchmark, we define two standardized splits across all environments: an \textit{easy split} with the short-horizon, symbolic-augmented instances, and a \textit{hard split} with the long-horizon, visual-only instances (Figure~\ref{fig:difficulty} right).
For diagnostic analysis, we additionally run matched ablations that vary one axis at a time: visual-only versus symbolic-augmented comparisons estimate the visual-abstraction bottleneck, while short- versus long-horizon comparisons under symbolic-augmented observations estimate the effect of planning horizon.
Together, these comparisons distinguish failures caused by constructing task-relevant spatial states from visual observations from failures caused by reasoning over action-conditioned state changes once the relevant state is explicit.

\textbf{Data Generation and Evaluation.}
All benchmark instances are generated using executable environment dynamics and reference solutions.
Each environment provides a gym-style interface for state reset, action execution, observation rendering, symbolic state extraction, and solution verification.
\DatasetName can synthesize arbitrarily many training examples, making the environments suitable for training agents with Reinforcement Learning.
\DatasetName supports two evaluation modes: single-shot evaluation, where the model predicts a complete plan from the initial observation, and multi-turn evaluation, where the model acts step by step using updated environment feedback.
We use task success rate as the primary metric, and further report state-parsing accuracy, step-wise success rate, and failure analyses to identify the sources of model errors.
\section{Evaluating Frontier MLLMs on \DatasetName}
\label{sec:experiment}

We evaluate performance on the \DatasetName benchmark using ten state-of-the-art MLLMs, including both open-weight and closed-weight models with advanced multimodal reasoning capabilities. Specifically, we assess seven representative open-weight MLLMs: Qwen3-VL-235B~\cite{bai2025qwen3}, Qwen3.5-397B~\cite{qwen3.5}, Qwen3.6-27B~\cite{qwen3.6-27b}, Llama-4-Scout~\citep{meta2025llama4}, GLM-4.6V~\cite{vteam2025glm45vglm41vthinkingversatilemultimodal}, Gemma-4-31B~\cite{team2026gemma}, and Kimi-K2.6~\cite{team2026kimi}, covering a broad range of model sizes from 27B to 1.1T.
In addition, we also include three state-of-the-art closed-weight models in our evaluation: Grok-4.3, GPT-5.4~\citep{gpt-5.4}, and Gemini-3.1-pro~\citep{gemini-3.1-pro}. 
To further understand the challenges of \DatasetName{}, we conduct a human study and report the average performance from experienced players as a reference. Details of the human evaluation protocol are provided in Appendix~\ref{sup:experiment_details}.

Our primary evaluation focuses on the agentic setting, where the models interact with the environment over multiple turns. We also report a single-shot ablation where models produce a complete plan from the initial input without intermediate environment feedback in Appendix~\ref{sup:additional_results}.
Through these evaluations, we ask two questions: 

\textit{\textbf{(i)}} How well do frontier MLLMs solve spatially grounded reasoning and planning tasks in \DatasetName{}?

\textit{\textbf{(ii)}} What is the major bottleneck of frontier models on tasks in \DatasetName{}?

We present the main results in Section~\ref{subsec:main_results}, provide a diagnostic probe in Section~\ref{subsec:diagnosis_failure}, and conduct further analysis in Section~\ref{subsec:visual_abstract}.
Full experimental configurations, prompts, and hyperparameters are provided in Appendix~\ref{sup:environments} and ~\ref{sup:experiment_details}. The average output tokens per turn for each model and environment are reported in Appendix~\ref{sup:additional_results}.

\subsection{Overall Results}
\label{subsec:main_results}
\begin{table}[b]
    \centering
    \vspace{-1.5em}
    \caption{Multi-turn evaluation success rate (\%) on \DatasetName. 
    Rows are ordered by average performance, with stronger models lower in the table.}
    \label{tab:online-results}
    {
    \begin{tabular}{lcccccc}
        \toprule
        & \multicolumn{2}{c}{Cube} & \multicolumn{2}{c}{Maze} & \multicolumn{2}{c}{Blox} \\
        \cmidrule(lr){2-3} \cmidrule(lr){4-5} \cmidrule(lr){6-7}
        Model & Easy & Hard & Easy & Hard & Easy & Hard \\
        \midrule
        Human & 86.6 & 85.0 & 82.5 & 78.8 & 78.6 & 62.5 \\
        \midrule
        \textit{Open-Weight models} \\
        Llama4-Scout  & 0.0 & 0.0 & 0.0 & 0.0 & 0.0 & 0.0 \\
        GLM-4.6V      & 4.0 & 0.0 & 2.5 & 0.0 & 0.0 & 0.0 \\
        Qwen3.6-27B   & 4.0 & 0.0 & 5.0 & 0.0 & 4.0 & 0.0 \\
        Qwen3-VL-235B & 29.6 & 0.0 & 16.3 & 0.6 & 5.3 & 0.0 \\
        Qwen3.5-397B  & 25.8 & 0.0 & 40.2 & 1.0 & 47.1 & 0.0 \\
        Gemma-4-31B   & 42.5 & 0.0 & 48.5 & 1.0 & 61.0 & 0.0 \\
        Kimi-K2.6      & \textbf{82.0} & 0.0 & \textbf{82.7} & \textbf{6.2} & \textbf{99.0}  & \textbf{2.2} \\
        \midrule
        \textit{Closed-weight models} \\
        Grok-4.3       & 46.7 & 0.0 & 43.8 & 0.0 & 25.0 & 0.0 \\
        Gemini-3.1-pro & 46.7 & 0.0 & 60.0 & 5.3 & 86.7 & \bf 8.3 \\
        GPT-5.4        &  \textbf{93.3} & 0.0 & \textbf{90.5} & \textbf{22.0} & \textbf{100.0} & 0.0 \\
        \bottomrule
    \end{tabular}
    }
    \label{tab:overall_results}
\end{table}

\begin{figure}[t]
    \centering
    \begin{minipage}[c]{0.59\linewidth}
        \centering
        \includegraphics[width=\linewidth]{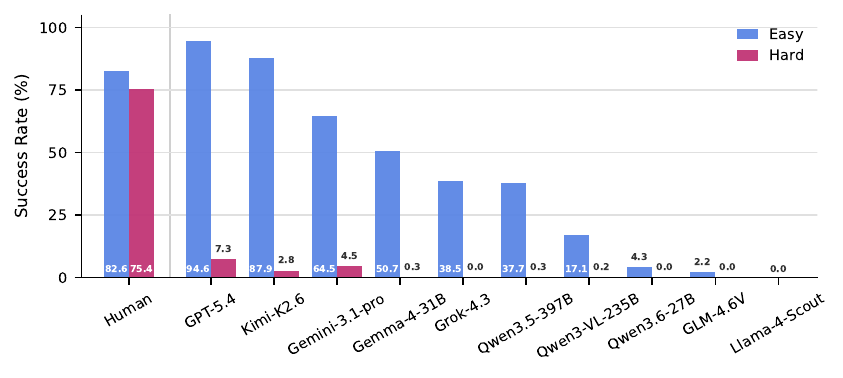}
    \end{minipage}%
    \hfill
    {\color{lightgray}\vrule width 0.4pt}%
    \hfill
    \begin{minipage}[c]{0.38\linewidth}
        \centering
        \includegraphics[width=\linewidth]{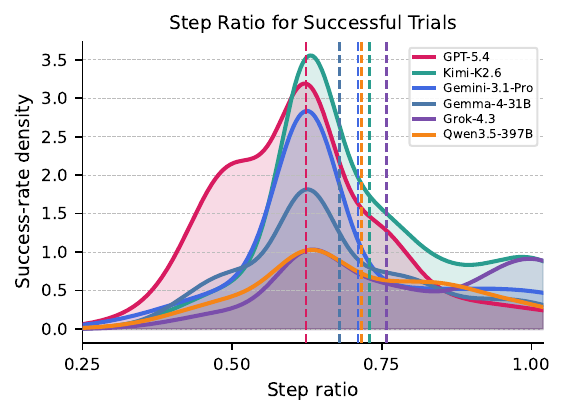}
    \end{minipage}
    \caption{
    \textbf{Multi-turn task success and interaction efficiency.}
    \textit{Left}: Average success rate across all the environments.
    \textit{Right}: Distribution of successful trials by the normalized step ratio (\textit{step / max\_step}); dashed lines mark each model’s mean.
}
    \vspace{-1.5em}
    \label{fig:step_ratio_success}
\end{figure}

\paragraph{Easy Settings Distinguish Model Capabilities.}
We present the results of all evaluated models in Table~\ref{tab:overall_results} and Figure~\ref{fig:step_ratio_success}. 
Overall, \DatasetName clearly differentiates current MLLMs in the \textit{easy} setting.
On easy instances, weaker open-weight models remain near-zero success rate, whereas frontier systems are capable of solving most of the tasks: GPT-5.4 reaches 93.3\%, 90.5\%, 100.0\% success rate on \Mcube, \Mmaze, and \Mbloxorz, respectively, surpassing the performance of experienced human players.
Among open-weight models, Kimi-K2.6 performs best, achieving 82.0\%, 82.7\%, and 99.0\% success rate across the three easy environments and substantially narrowing the gap to closed-weight models.
Gemma-4-31B is also notable among smaller models, surpassing the larger Qwen3-VL-235B and Qwen3.5-397B models across all three easy settings and offering a strong size-performance trade-off.
Figure~\ref{fig:step_ratio_success} further visualizes the average step ratio (executed steps / max steps) among successful trajectories. The results show that GPT-5.4 is the most efficient problem solver, achieving the highest success rates while using the smallest fraction of the available interaction budget. 
In contrast, several strong models, such as Kimi-K2.6 and Grok-4.3, require substantially more steps to solve the same tasks, suggesting that their success in the easy setting may rely more on direct search than on efficient spatial planning.
Together, these results show that the easy split of \DatasetName provides a reliable way to evaluate MLLMs' spatial reasoning and planning abilities in terms of both task success and solution efficiency.

\paragraph{Hard Settings Expose Common Failures of Frontier Models.}
Compared with the easy split, the hard setting remains largely unsolved for frontier MLLMs: all models score 0.0\% on \Mcube, while the best success rates are only 22.0\% on \Mmaze from GPT-5.4 and 8.3\% on \Mbloxorz from Gemini-3.1-pro.
This sharp degradation from easy to hard stands in contrast to human performance: experienced players show a drop of only 1.6–16.1 percentage points when moving from the easy to the hard setting, whereas frontier MLLMs suffer substantially larger declines.
These results suggest that current frontier MLLMs remain brittle on complex spatially grounded reasoning and planning tasks.
We next provide a detailed analysis of their failure modes in the hard setting.

\subsection{Visual Abstraction is the Major Source of Failure}
\label{subsec:diagnosis_failure}

\begin{figure}[b]
    \centering
    \begin{minipage}[c]{0.48\linewidth}
        \centering
        \includegraphics[width=\linewidth]{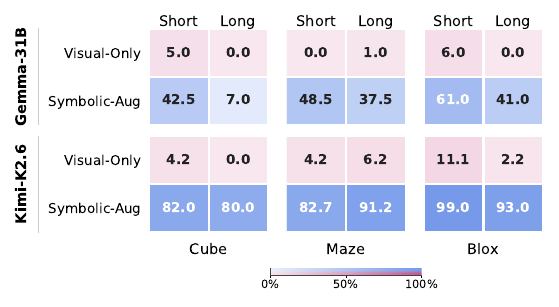}
    \end{minipage}%
    \hfill
    {\color{lightgray}\vrule width 0.4pt}%
    \hfill
    \begin{minipage}[c]{0.24\linewidth}
        \centering
        \includegraphics[width=\linewidth]{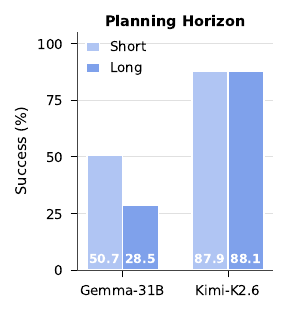}
    \end{minipage}
    \begin{minipage}[c]{0.24\linewidth}
        \centering
        \includegraphics[width=\linewidth]{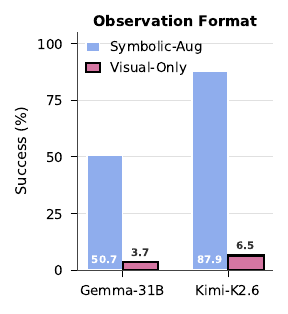}
    \end{minipage}
    \caption{
    \textbf{Planning Horizon vs.\ Observation Format.}
    \textit{Left}: Success rates for the $2\times2$ diagnostic crossing planning horizon (short/long) and observation format (symbolic-augmented/visual-only) across \Mcube{}, \Mmaze{}, and \Mbloxorz{}. Color saturation encodes success rate.
    \textit{Right}: Environment-averaged comparisons isolating the impact of planning horizon and observation format.
}
    \label{fig:depth_abstraction_ablation}
\end{figure}

The hard split combines two difficulty axes: longer planning horizons and visual-only observations without symbolic states.
To disentangle their effects, we evaluate a $2\times2$ diagnostic design over planning horizon (short versus long) and observation format (symbolic-augmented versus visual-only) on two of the best open-weight models: Gemma-4-31B and Kimi-K2.6.
Figure~\ref{fig:depth_abstraction_ablation} reports their resulting success rates across \Mcube{}, \Mmaze{}, and \Mbloxorz{}.
The $2\times2$ conditional panels show all four settings for each model-environment pair, while the summary panels average over environments to isolate the effects of planning horizon and observation format under controlled comparisons.

\textbf{Planning Horizon Mainly Affects Weaker Models.}
When symbolic state is available, the effect of increasing the planning horizon is model-dependent.
Gemma-4-31B drops on all three environments, from 42.5\% to 7.0\% on \Mcube{}, from 48.5\% to 37.5\% on \Mmaze{}, and from 61.0\% to 41.0\% on \Mbloxorz{}.
This corresponds to a 22.2-percentage point decline averaged over all the environments.
By contrast, Kimi-K2.6 remains nearly unchanged on the same comparison, with average success moving from 87.9\% to 88.1\%.
Thus, longer planning chains remain challenging for weaker models like Gemma-4-31B, but are not the dominant source of failure for stronger frontier models such as Kimi-K2.6.

\textbf{Observation Format Reveals the Major Bottleneck.}
In contrast to planning horizon, observation format induces a much larger and more consistent degradation across models.
At the matched short horizon, removing symbolic state reduces Gemma-4-31B from 50.7\% to 3.7\% success rate on average, and reduces Kimi-K2.6 from 87.9\% to 6.5\%.
The performance collapse occurs in every environment: short-horizon visual-only success remains at 5.0\%, 0.0\%, and 6.0\% for Gemma-4-31B, and at 4.2\%, 4.2\%, and 11.1\% for Kimi-K2.6 across \Mcube{}, \Mmaze{}, and \Mbloxorz{}.
The comparison is especially clear for Kimi-K2.6: the model is almost insensitive to longer symbolic-augmented planning, yet fails when it must infer the underlying spatial state directly from visual observations.
Therefore, compared with long-horizon planning, constructing reliable spatial states before planning is the major bottleneck for frontier MLLMs.

\begin{figure}[t]
    \vspace{-3.0em}
    \centering
    \begin{minipage}[c]{0.48\linewidth}
        \centering
        \includegraphics[width=\linewidth]{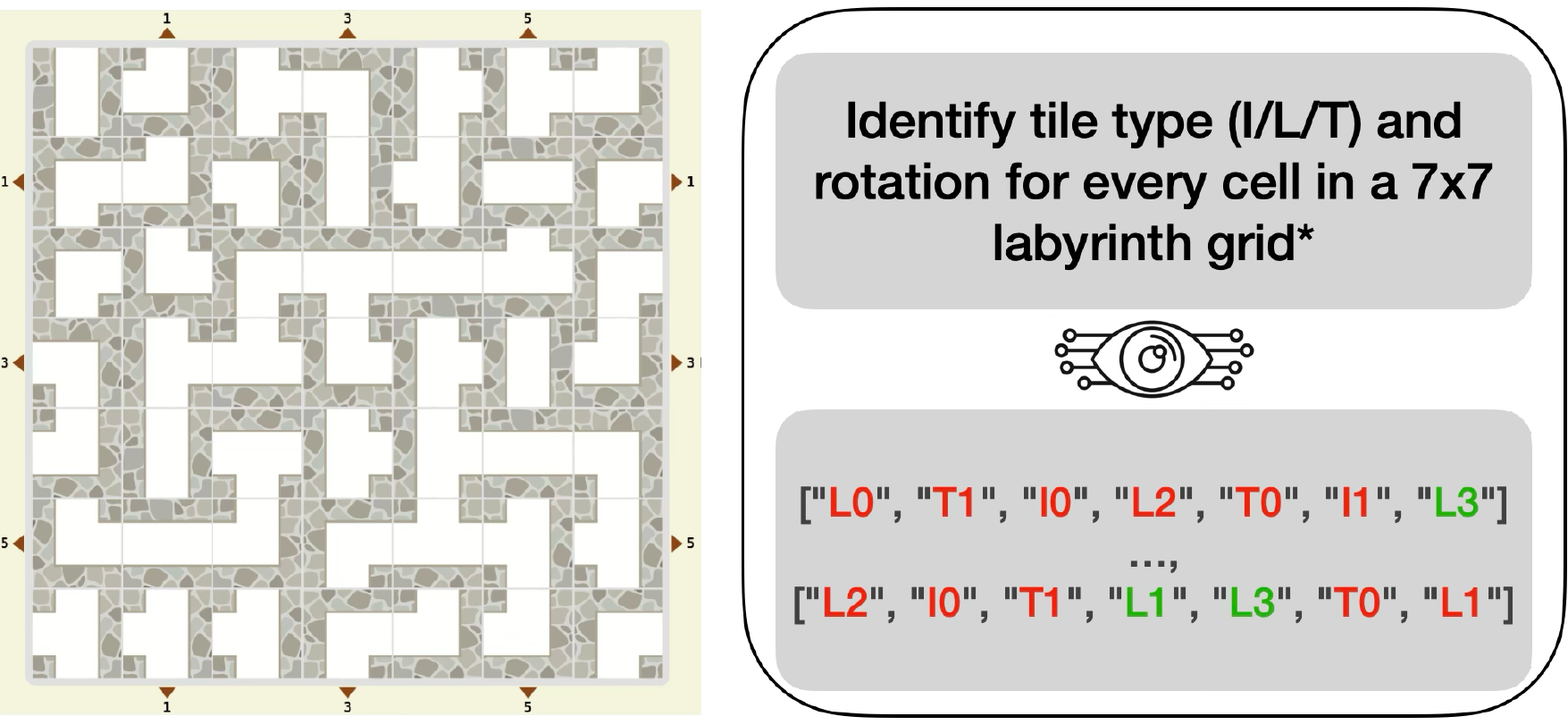}
    \end{minipage}%
    \hfill
    \begin{minipage}[c]{0.015\linewidth}
        \centering
        {\color{lightgray}\rule{0.1pt}{0.15\textheight}}
    \end{minipage}
    \hfill
    \begin{minipage}[c]{0.48\linewidth}
        \centering
        \includegraphics[width=\linewidth]{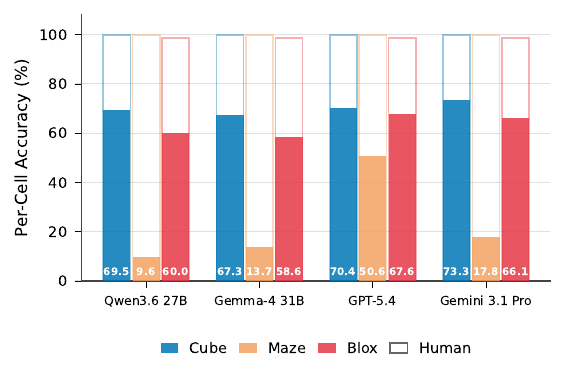}
    \end{minipage}
     \vspace{-1.0em}
    \caption{
    \textbf{Spatial State Parsing}. \textit{Left}: Example of the \Mmaze parsing task, where models extract the tile type (I/L/T) and rotation (0-3) of each cell from a rendered 7x7 board (see Appendix~\ref{sup:experiment_details} for \Mcube{} and \Mbloxorz{} tasks). \textit{Right}: Per-cell parsing accuracy across \Mcube, \Mmaze and \Mbloxorz. Outlined bars indicate human performance ($\geq 98\%$ on all tasks), highlighting the substantial gap between MLLMs and humans in visual-to-spatial abstraction.
    }
    \vspace{-1.0em}
    \label{fig:perception_results}
\end{figure}

\subsection{Frontier Models Fail to Construct Visual Abstraction Accurately}
\label{subsec:visual_abstract}

\begin{wrapfigure}{r}{0.45\linewidth}
    \vspace{-1.0em}
    \centering
    \includegraphics[width=0.95\linewidth]{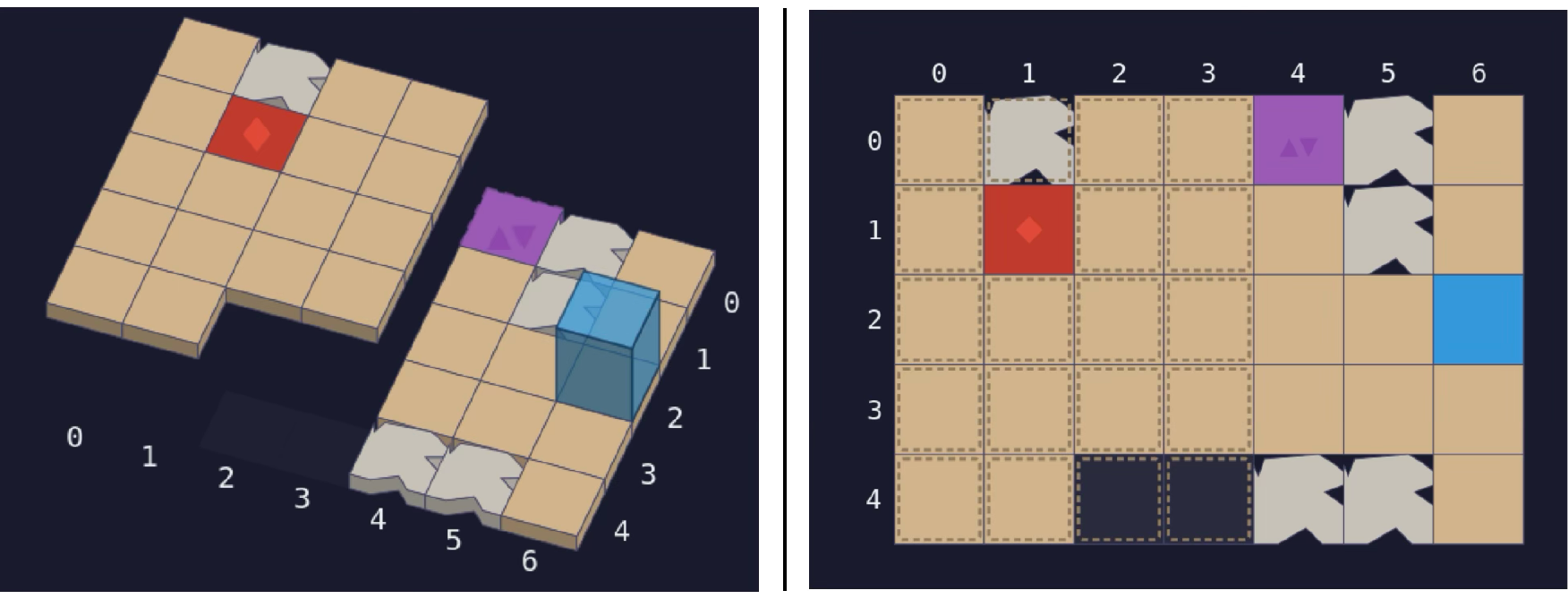}\\[-0.4em]
    \makebox[\linewidth]{\fontsize{7}{8.5}\sffamily\fontseries{sb}\selectfont\makebox[0.5\linewidth]{(a) Isometric 3D}\makebox[0.5\linewidth]{(b) Flat 2D}}\\[-0.4em]
    {\color{lightgray}\rule{\linewidth}{0.1pt}}\\[0.05em]
    \includegraphics[width=\linewidth]{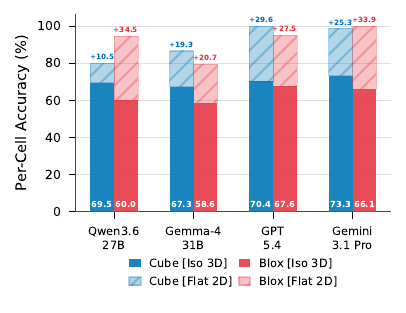}
    \vspace{-1.5em}
    \caption{{\textbf{Isometric Perception Gap.}} \textit{Top}: Same \Mbloxorz{} board rendered in isometric 3D (left) and flat 2D (right). \textit{Bottom}: Per-cell accuracy on \Mcube{} and \Mbloxorz{} under both viewpoints. All models improve substantially under the flat view.}
    \label{fig:perspective_gap}
    \vspace{-3.0em}
\end{wrapfigure}

The diagnostic results of Section~\ref{subsec:diagnosis_failure} show that removing symbolic states collapses performance far more than increasing the planning horizon, confirming that the primary bottleneck lies in visual-to-spatial abstraction rather than in planning itself. To understand why visual abstraction fails, we designed three \textbf{spatial state parsing} tasks (Figure~\ref{fig:perception_results}) that isolate state estimation from downstream planning: given a rendered observation of the environment, the model must extract a structured spatial representation without any planning or action (more details in the Appendix~\ref{sup:experiment_details}). In \Mcube, the model extracts the filled/empty cell pattern of a $5 \times 5$ grid from a three-dimensional isometric rendering. In \Mmaze, the model must discriminate which edges of each tile are corridors versus walls to infer tile type (I/L/T) and rotation ($0$--$3$) across all 49 cells of a $7 \times 7$ labyrinth, a task that combines fine-grained spatial discrimination with dense structured extraction. Finally in \Mbloxorz, the task is to recover the grid layout, block position and orientation, and goal location from a board rendering. 

\paragraph{MLLMs Have Weak Spatial Perception.} 
The results of spatial state parsing are presented in Figure~\ref{fig:perception_results}. 
Across all the environments, frontier models largely fall behind the human performance.
For example, GPT-5.4 achieves only 70.4\%, 50.6\%, and 67.6\% per-cell accuracy on \Mcube{}, \Mmaze{}, and \Mbloxorz{}, respectively. In comparison, human participants achieve near-perfect accuracy on all three tasks, indicating that spatial state parsing is straightforward for humans.
To isolate the source of this failure, we further ablate the effect of viewpoint by comparing 3D isometric and flat 2D top-down renderings of the same state as illustrated in Figure~\ref{fig:perspective_gap}. 
The results indicate that all the models obtain much better performance under flat 2D view, with performance rising from $67$--$73\%$ to $80$--$99\%$ for \Mcube, and from {$59$--$68\%$} to {$79$--$100\%$} for \Mbloxorz.
This \textbf{isometric perception gap} is consistent across open-weight and closed-weight models, suggesting that the isometric projection is the primary source of perceptual failure. In contrast, human participants perform equally well on both viewpoints.

\begin{figure}[t]
    \centering
    \includegraphics[width=\linewidth]{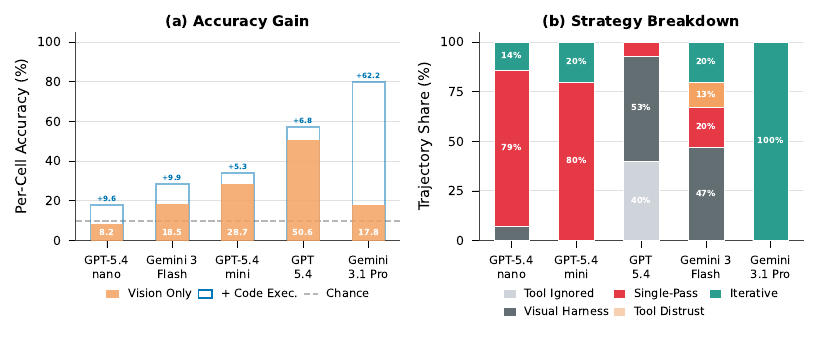}
    \vspace{-2.0em}
    \caption{\textbf{Agentic Vision on \Mmaze{} Board Parsing.} \textit{Left}: Per-cell parsing accuracy with vision only (filled bars) and with code execution (outlined bars). Annotations indicate accuracy gains in percentage points, with Gemini 3.1 Pro achieving the largest improvement (+62.2 pp). \textit{Right}: Distribution of tool-use strategies across trajectories. Gemini 3.1 Pro consistently adopts iterative parsing, whereas other models predominantly rely on single-pass parsing or less systematic tool use.
    }
    \label{fig:agentic_vision}
    \vspace{-1.0em}
\end{figure}

\paragraph{Visual Perception with Agentic Tool Use.}
Prior work has shown that current MLLMs can leverage external tools, such as code execution, to improve multimodal reasoning and understanding~\cite{zheng2025deepeyes}. 
To examine whether agentic tool use can help mitigate perceptual bottlenecks in spatial state parsing, we further evaluate MLLMs on these tasks under a tool-use setting. 
The results are shown in Figure~\ref{fig:agentic_vision} (a). Code execution consistently improves MAZE board parsing accuracy, although the gains vary substantially across models. Smaller models such as GPT-5.4 nano and GPT-5.4 mini achieve only modest improvements of 5 to 10 percentage points, whereas Gemini 3.1 Pro improves from 17.8\% to 80.0\%, achieving the largest gain of 62.2 percentage points.
To better understand these differences, we classify individual trajectories into five tool-use strategies: \textcolor{stratIgnored}{\textbf{Tool Ignored}}, where the model does not invoke code execution; \textcolor{stratHarness}{\textbf{Visual Harness}}, where code is used to render visual aids but classification remains primarily visual; \textcolor{stratSingle}{\textbf{Single-Pass Parsing}}, where a single script extracts the full board state; \textcolor{stratDistrust}{\textbf{Tool Distrust}}, where the model overrides code outputs based on visual inspection; and \textcolor{stratIterative}{\textbf{Iterative Parsing}}, where the model repeatedly crops tiles, samples edge pixels, and refines thresholds over multiple rounds. 
The results are illustrated in Figure~\ref{fig:agentic_vision} (b). Notably, Gemini 3.1 Pro is the only model that consistently adopts iterative parsing, suggesting that systematic and iterative tool use can substantially improve visual-to-spatial abstraction.

\section{Error Analysis}
\label{sec:spatial_world_modeling}
Section~\ref{subsec:diagnosis_failure} demonstrates that visual abstraction is the primary bottleneck for frontier models to solve \DatasetName{}. We further quantify the remaining errors in the easy setting, where symbolic states are available, by categorizing failed multi-turn trajectories into four distinct types:

\begin{list}{\arabic{enumi}.}{\usecounter{enumi}
    \setlength{\leftmargin}{1.45em}
    \setlength{\labelwidth}{1.0em}
    \setlength{\labelsep}{0.35em}
    \setlength{\itemsep}{0.05em}
    \setlength{\parsep}{0pt}
    \setlength{\topsep}{0.05em}}
    \item \textcolor{errInvalid}{\textbf{Invalid Action}}, where the model emits an illegal or unparsable action;
    \item \textcolor{errMechanics}{\textbf{Spatial Mechanics}}, where the model incorrectly simulates task constraints or state transitions;
    \item \textcolor{errFeedback}{\textbf{Feedback Use}}, where the model fails to incorporate intermediate feedback;
    \item \textcolor{errSearch}{\textbf{Inefficient Search}}, where the model spends the finite interaction budget on circular, redundant, or low-value exploration.
\end{list}

\begin{figure}[h]
    \centering
   
    \includegraphics[width=0.99\linewidth]{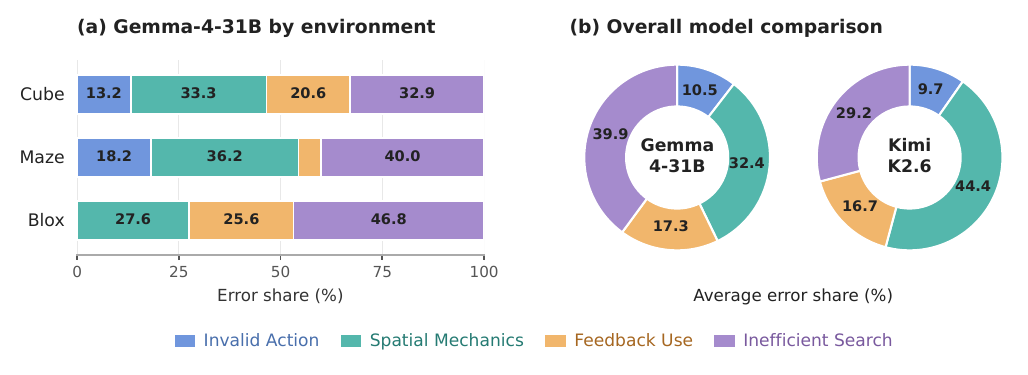}
    \vspace{-0.5em}
    \caption{\textbf{Error Categorization for Failed Multi-Turn Trajectories.}
    \textit{Left}: error breakdown for failed Gemma-4-31B easy-setting trajectories with symbolic state feedback. When multiple error types apply to one trajectory, we assign fractional credit to each label. \textit{Right}: overall error distribution for Gemma-4-31B and Kimi-K2.6, averaged across the three environments.}
    \label{fig:error_categorization}
\end{figure}

Figure~\ref{fig:error_categorization} shows the breakdown for Gemma-4-31B and Kimi-K2.6. For Gemma-4-31B, the dominant errors are inefficient search ($39.9\%$ averaged over environments) followed by incorrect spatial mechanics simulation ($32.4\%$). 
Among Kimi's failures, spatial mechanics is the dominant category ($44.4\%$), while inefficient search accounts for a smaller share ($29.2\%$). Invalid-action and feedback-use errors occur in similar proportions across the two models.
This pattern is consistent with the planning-horizon ablation in Section~\ref{subsec:diagnosis_failure}, where Kimi maintains high success rates under longer planning horizons when symbolic states are available. We hypothesize that Kimi's stronger planning capabilities reduce failures caused by inefficient search, leaving spatial-transition modeling as a more prominent challenge among its remaining failures.
The results also indicate that spatial mechanics errors are more diagnostic for \DatasetName, which directly reflect whether the model can simulate task-specific spatial constraints and state transitions, a prerequisite for solving these tasks when symbolic feedback is available.
These failures may be improved by explicit state tracking, transition-consistency checks, and search strategies that prune repeated or invalid branches.

\section{Conclusion}
\label{sec:conclusion}

This paper introduced \DatasetName, a multimodal benchmark for spatially grounded reasoning and planning. \DatasetName provides a diagnostic testbed that isolates failures in visual-to-spatial abstraction from failures in long-horizon action-conditioned reasoning in agentic environments. Our empirical evaluation reveals that state-of-the-art MLLMs still struggle substantially on \DatasetName, primarily due to bottlenecks in visual-to-spatial abstraction. 
\DatasetName fills a critical gap in multimodal reasoning evaluation by shifting the focus from outcome accuracy to process-oriented, multi-step reasoning that requires coherent visual abstraction and spatial understanding. 
By systematically exposing where perception, abstraction, and planning break down, \DatasetName provides a foundation for measuring progress toward more reliable multimodal agents. We hope that \DatasetName will elicit further progress and innovation in MLLM development, unlocking stronger reasoning and planning abilities in complex multimodal environments—capabilities that are essential for real-world, embodied, and general-purpose intelligence.

\textbf{Limitations}.
\DatasetName investigates spatially grounded reasoning in synthetic, game-inspired environments. This design supports controlled analysis of visual-to-spatial abstraction and long-horizon planning. However, the tasks in the current environment do not reflect the full richness of real-world embodied interaction. In particular, the visual-abstraction tasks in \DatasetName are based on discrete, rule-governed spatial layouts with explicit state variables and action dynamics. Real-world settings, by contrast, often involve richer geometry, object affordances, partial observability, noisy perception, continuous control, and underspecified goals. Our evaluation also relies on a relatively simple agent loop and does not systematically examine more advanced agentic mechanisms, such as persistent memory or interleaved reasoning with images. We therefore position \DatasetName as a diagnostic testbed for targeted spatial world-modeling capabilities, rather than as a comprehensive measure of general embodied multimodal intelligence.

\section*{Broader Impact}
\DatasetName can support the development of more reliable multimodal agents by exposing failures in spatial state construction, transition modeling, planning, and feedback use. Because it uses synthetic puzzle environments and no private or sensitive data, its direct societal risks are limited. However, progress on spatially grounded reasoning may inform future autonomous systems, where failures could have safety consequences. Thus, strong performance on \DatasetName should not be viewed as evidence of safe real-world autonomy, and should be complemented by evaluations of robustness, uncertainty, and deployment context.

\section*{Acknowledgements}

We thank Yueqing Shen, Shuo Wen, Jeremy Goumaz, Allen Zhao, Jiwoong Sohn and Mu Shen for their valuable suggestions, which helped improve the clarity of the manuscript.
We gratefully acknowledge the support of the Swiss National Science Foundation (SNSF) starting grant TMSGI2\_226252/1, SNSF grant IC00I0\_231922, SNSF grant 10.004.411, Swiss Cancer League KFS-6340-02-2025-R, Kamprad Foundation (Cancer Immunotherapy Networking Program) and the Swiss AI Initiative (\#ab006, \#ab0023, \#a0174). M.B. is a CIFAR Fellow in the Multiscale Human Program.

\bibliographystyle{plain}
\bibliography{ref}

\newpage
\appendix

\clearpage
\section{Environments}
\label{sup:environments}

\subsection{Task description}

\textbf{\Mcube} is a 3D spatial assembly puzzle inspired by Happy Cube. Each instance presents six jigsaw-style pieces corresponding to the faces of a $5 \times 5 \times 5$ cube. Each piece is represented as a $5 \times 5$ grid with a fixed central $3 \times 3$ region, while the boundary cells define bump-and-gap patterns. The goal is to assemble all six pieces into a closed cube whose adjacent edges align without gaps or overlaps. Solving the task requires recovering each piece's geometry from the observation, assigning every piece to a distinct cube face, and choosing a valid orientation, equivalently a rotation and optional flip specified by the left and top edge numbers in the local face view. The task therefore combines local edge matching with global corner consistency, while requiring the model to search over face assignments, orientations, and whole-cube rotational symmetries.

\textbf{\Mmaze} is a 2D spatial-planning puzzle inspired by \textit{The aMAZEing Labyrinth}. Each state contains a $7 \times 7$ maze and one off-board spare tile. Tiles have I-, L-, or T-shaped corridors with discrete orientations, and the agent must move the player to the target by deciding when to reshape the board and when to navigate through connected corridors. At each decision step, the agent may either insert the spare tile into a legal row or column with a chosen rotation, thereby shifting the board and ejecting the opposite tile, or move the player to a reachable cell through connected corridors. Because insertions can change maze connectivity, spare tile identity, player position, and object positions, the task tests visual parsing of tile structure, state tracking under dynamic topology, and multi-step planning over action-induced board transformations.

\textbf{\Mbloxorz} is a 3D navigation environment with pose-dependent dynamics. It features a $1 \times 1 \times 2$ rectangular block that rolls across a platform grid using cardinal actions \texttt{UP}, \texttt{DOWN}, \texttt{LEFT}, and \texttt{RIGHT}. The block may stand upright or lie horizontally, and its pose determines which cells it occupies and which transitions are valid: an upright block tips over, while a horizontal block may roll upright or shift while remaining horizontal. The goal is to reach the target configuration, often requiring the block to stand upright on a goal cell, while invalid moves can cause the block to leave the platform or become unsupported. Some boards also include elevators and crumble blocks, which introduce non-uniform and history-dependent terrain dynamics. This task emphasizes spatial state tracking, reasoning over pose-dependent and tile-dependent transitions, and long-horizon consequence prediction, since a move that is safe in one pose or earlier state may be invalid or strategically harmful in another.

\subsection{Dataset Generation}

\paragraph{\Mcube{}.}
Each puzzle instance is constructed by (1) selecting a valid cube solution from a library of pre-computed assemblies, (2) decomposing it into six $5 \times 5$ piece grids, and (3) rendering the pieces as 3D isometric images with randomized viewpoints. The resulting dataset contains puzzles of varying difficulty, controlled by the number of missing pieces the model must place and whether piece flips are allowed.

\paragraph{\Mmaze{}.}
For each seeded board and starting position, we solve a forward shortest-turns problem to compute the minimal depth $D$ (number of Insert+Move turns to reach the target). Instances are generated across multiple depths, with $D$ serving as a lower bound and proxy for difficulty. To ensure diversity, at most one solution per (board, seed, depth) triplet is retained. The evaluation uses the initial board state of each solved instance, allowing for multiple valid solutions.

\paragraph{\Mbloxorz{}.}
Levels are generated procedurally via reverse BFS from the goal cell. Starting from a block standing upright on the goal, we expand all reachable states backwards through cardinal roll actions to compute the shortest solution depth for every possible starting configuration on a full grid. A starting state is then sampled at the target depth, and cells not required by any optimal path are carved out as void with controlled probability, producing sparse platforms that constrain the solution space. 
The easy split uses a compact 5 × 7 grid with one elevator and crumble tiles (target depth 10), while the hard split uses the same grid size, elevator count, and tile types but requires longer solutions (target depth 16).
Each generated level is verified by forward BFS to ensure the optimal solution matches the target depth exactly.

\newpage
\subsection{Task Example}
\label{app:task_example}

\begin{figure}[h]
    \centering
    \includegraphics[width=0.99\linewidth]{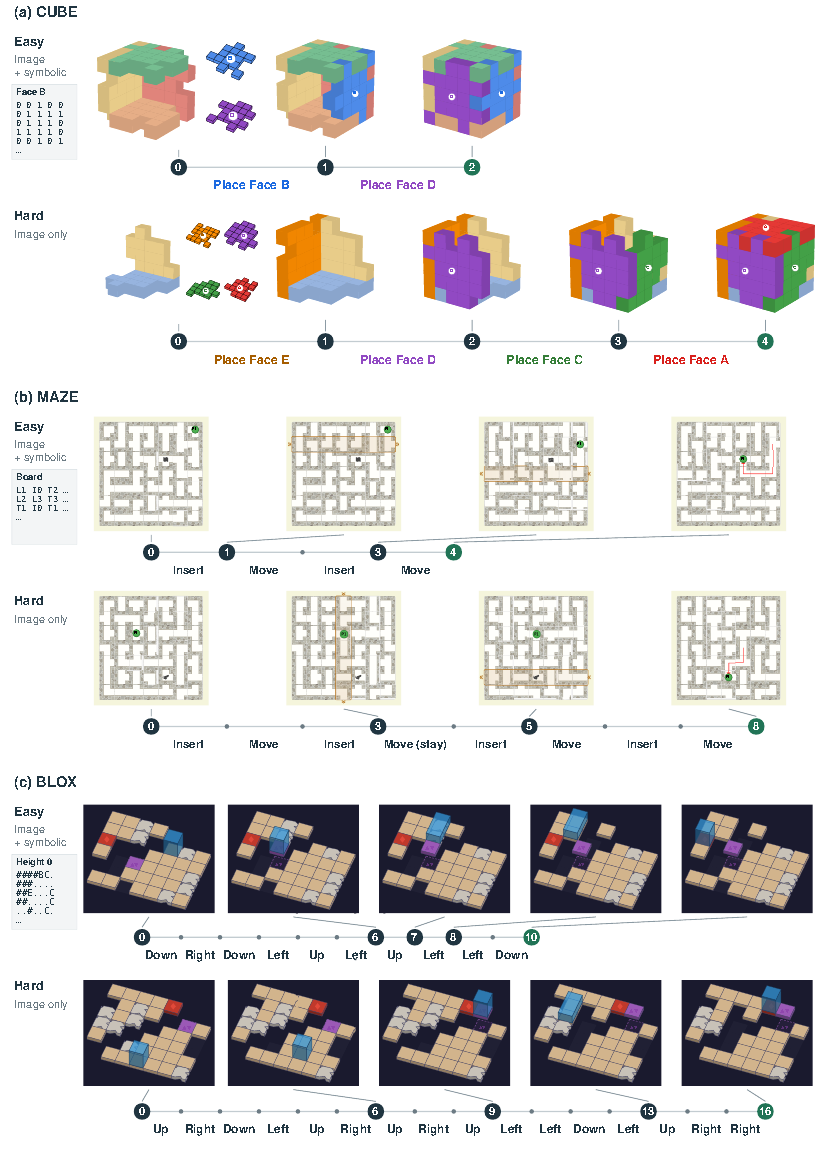}
    \caption{Illustration of easy and hard task examples of \Mcube, \Mmaze and \Mbloxorz.}
    \label{fig:task_illustration}
\end{figure}

Figure~\ref{fig:task_illustration} presents representative easy and hard tasks from \Mcube, \Mmaze and \Mbloxorz with valid reference trajectories. Easy inputs combine images with symbolic state representations, whereas hard inputs omit the additional symbolic representation. 

\begin{table}[t]
    \centering
    \small
    \caption{
        Environment configurations for the easy and hard splits. Rendered images are provided in all settings. Symbolic annotations provide an explicit description of the spatial state.        
    }
    \label{tab:difficulty_configs}
    \vspace{0.5em}
    \begin{tabular}{lllcc}
        \toprule
        Environment & Split & Planning horizon
        & Symbolic annotations & Action budget \\
        \midrule
        \Mcube & Easy & 2 missing pieces & \cmark & 4 \\
             & Hard & 4 missing pieces & \xmark  & 8 \\
        \midrule
        \Mmaze & Easy & 2 insert--move turns & \cmark & 6 \\
             & Hard & 4 insert--move turns & \xmark  & 12 \\
        \midrule
        \Mbloxorz & Easy & 10 moves & \cmark & 16 \\
             & Hard &  16 moves & \xmark  & 24 \\
        \bottomrule
    \end{tabular}
\end{table}

\newpage
\subsection{Prompts}
\tcbset{
    recapprompt/.style={
        enhanced standard,
        breakable,
        colback=white,
        colframe=black,
        colbacktitle=black,
        coltitle=white,
        fonttitle=\bfseries,
        boxrule=0.8pt,
        arc=2pt,
        left=10pt,
        right=10pt,
        top=8pt,
        bottom=8pt,
        before skip=8pt,
        after skip=12pt,
        notitle after break,
        toprule at break=2mm,
        bottomrule at break=0.8pt
    }
}
\DefineVerbatimEnvironment{PromptCode}{Verbatim}{
    fontsize=\small,
    breaklines=true,
    breakanywhere=true,
    breaksymbolleft={}
}


\begin{tcolorbox}[recapprompt,title={Prompt for \Mcube}]
\raggedright
\setlength{\parindent}{0pt}
\setlength{\parskip}{0.3em}

You are a spatial-reasoning assistant that solves Happy Cube puzzles.

\textbf{INPUT}

\textbullet{} Six jigsaw-style pieces, labelled A--F.\\
\textbullet{} Each piece is described by the bump/gap pattern on its four edges (Edge1, Edge2, Edge3 and Edge4) as shown in the image, the center squares are always filled.

\textbf{TASK}

Build a \(5 \times 5 \times 5\) cube that uses every piece exactly once.

\textbullet{} Assign each piece to one cube face (Top, Bottom, Front, Back, Left, Right) with proper orientation (rotation and/or flip).\\
\textbullet{} The finished cube closed perfectly---no overlaps, no gaps.

The first and last 'square' of the edge contribute to the corners of the cube, make sure:

\textbullet{} When two cube faces touch, the non-corner part of their edges must be complementary bit-for-bit.\\
\textbullet{} At cube corners (where three faces meet), all connecting edges must align seamlessly.

\textbf{NOTE}

For each piece, report the orientation via the following mental exercise:

1. Hold the finished cube in both hands.\\
2. Turn the whole cube until the face you are about to name is now facing you like the front of a box.\\
3. In this view, write down the edge numbers located on left and top.

\textbullet{} Format: \verb|`<piece>: (<face>, <left_edge>, <top_edge>)`|

Note that left-edge and top-edge must be adjacent.

Below is the 6 faces where 0 = gap and 1 = bump.

\begin{PromptCode}
FACE A: {array}
FACE B: {array}
FACE C: {array}
FACE D: {array}
FACE E: {array}
FACE F: {array}
\end{PromptCode}

The following pieces are already correctly placed and LOCKED:

\begin{PromptCode}
```
B: (Front, 4, 3)
C: (Top, 3, 2)
D: (Left, 1, 2)
F: (Back, 4, 1)
```
\end{PromptCode}

\textbf{OUTPUT}

- This is an interactive evaluation. Think step by step and place exactly ONE non-locked piece at each turn.\\
- If you place a piece onto a face occupied by another non-locked piece, the previous occupant is replaced.\\
- Locked pieces cannot be replaced.\\
- Your response should end with:

\begin{PromptCode}
```
<piece>: (<face>, <left_edge>, <top_edge>)
```
\end{PromptCode}

\textbf{CURRENT STATE}

- You need to place: A, E.\\
- You have at most 4 turns.\\
- After each placement you will receive feedback: updated 3D views of the cube, a list of seam conflicts (if any).\\
- The feedback will show an updated 3D view of the current cube state.\\
- You may re-place a non-locked piece on a different non-locked face; if that face already has another non-locked piece, the previous placement will be evicted.

\end{tcolorbox}

\clearpage
\begin{tcolorbox}[recapprompt,title={Prompt for \Mmaze}]
\raggedright
\setlength{\parindent}{0pt}
\setlength{\parskip}{0.3em}

You are a spatial-planning assistant that plays Labyrinth from arbitrary board states.

\textbf{INPUT}

You will be given the current game observation:

- Board (7x7): each tile has corridors/walls (shape in \{I, L, T\} plus rotation in \{0,1,2,3\}), indexed 0..6 on both axes.\\
- Objects grid (7x7): a Python list of lists of strings (letters or '-' for empty) indicating item placement.\\
- Extra tile: the tile currently out of the board (with shape/rotation), encoded by a token (see conventions below).\\
- Player position: (x,y) of our player, 0-based.\\
- Target: a treasure object represented by its name mapped to a letter (A-X) in symbolic representation.\\
- Visual encoding: the player is a green circular pawn labeled ``P1''; the target is a black icon of the object.\\
- Right-side panel on the image: shows ``Extra Tile'' and ``Target'' sections with the current extra tile and the target.\\
- last\_insert: the previous insertion, if any, as (side, slot); used to enforce the ``no immediate reversal'' rule.

\textbf{RULES}

\textbf{INSERT action:}

- Set the loose tile to the specified rotation (see tile conventions below), then insert it from side into slot.\\
- This shifts that entire row/column by one; the opposite edge tile is ejected and becomes the new loose tile.\\
- Direction of ejection: Left inserts eject Right edge; Right ejects Left; Top ejects Bottom; Bottom ejects Top.\\
- Any pawns on that line shift with the tiles; if pushed off, they wrap to the newly inserted tile on the opposite edge.\\
- Objects behavior: treasures/items stay attached to their tiles; they shift with the row/column and if on the ejected tile, they leave the board (no wrap) and only re-enter if that tile is later re-inserted.\\
- No immediate reversal: you cannot insert from the opposite side into the same slot as last\_insert.

\textbf{MOVE action:}

- Move your pawn along open corridors to any reachable cell (unlimited distance along connected paths).\\
- Adjacency rule: a step to an orthogonal neighbor is legal only if both tiles open toward each other (e.g. W/E).\\
- To stay, output MOVE(); the keyword is required (not optional) when using plain-text action formats.\\
- Success is declared if the Target is reached after moving.

At each step, freely choose INSERT or MOVE -- there is no mandatory ordering between them.

\textbf{Tile/token convention for Extra}

- Shapes: I (straight corridor), L (corner/bend), T (three-way junction).\\
- Openings are on North (up), East (right), South (down), West (left).\\
- Rotation is clockwise; board top is North.

\textbf{Visual Descriptions:}

- I (straight): A corridor connecting two opposite sides\\
\quad * I0 = vertical corridor (connects top-bottom, North-South)\\
\quad * I1 = horizontal corridor (connects left-right, East-West)\\
\quad * I2 == I0, I3 == I1 (same visual appearance). Prefer I0/I1.

- L (corner): A bent corridor connecting two adjacent sides\\
\quad * L0 = connects top+right (North+East)\\
\quad * L1 = connects right+bottom (East+South)\\
\quad * L2 = connects bottom+left (South+West)\\
\quad * L3 = connects left+top (West+North)

- T (junction): A corridor with three openings\\
\quad * T0 = opens to top+left+right (North+East+West)\\
\quad * T1 = opens to right+bottom+top (East+South+North)\\
\quad * T2 = opens to bottom+left+right (South+West+East)\\
\quad * T3 = opens to left+top+bottom (West+North+South)

\textbf{NOTE}

- Grid coordinates: x,y in [0..6]; x is column index (left-right), y is row index (top-bottom).\\
- Inserts:\\
\quad - side in \{left, right, top, bottom\}.\\
\quad - slot in \{1,3,5\} (the only shove-able rows/columns).\\
\quad - For left/right, slot = row index (y). For top/bottom, slot = column index (x).\\
- rotation in \{0,1,2,3\} = absolute rotation (0, 90, 180, 270 degrees clockwise) to set the loose tile before insertion following the convention above.\\
Do not apply relative rotations; ignore the current extra tile orientation.\\
- The target is a single uppercase letter (A-X).\\
- The extra tile token encodes shape+rotation: I0/I1, L0..L3, T0..T3.\\
- Do not ask for more information; use only the provided image/text to decide.

\textbf{TASK}

Choose exactly one action -- either INSERT or MOVE -- to progress toward the Target (reach it if possible; otherwise improve your position for future steps).

\textbf{ACTION SPACE (exact output strings)}

- INSERT:
\begin{PromptCode}
Action: INSERT(side='<left|right|top|bottom>', slot=<1|3|5>, rotation=<0|1|2|3>)
\end{PromptCode}
- MOVE:
\begin{PromptCode}
Action: MOVE(x=<0..6>, y=<0..6>) (use MOVE() to stay)
\end{PromptCode}
\textbf{OUTPUT}

- Your chosen action must be legal and obey the rules above.\\
- Your response should end with:

\begin{PromptCode}
```
Action: INSERT(side='<left|right|top|bottom>', slot=<1|3|5>, rotation=<0|1|2|3>)
```
or
```
Action: MOVE(x=<0..6>, y=<0..6>)
```
\end{PromptCode}

\textbf{STATE}
\begin{PromptCode}
Last Insert: None
Board grid: {array}
Objects grid: {array}
Target: Treasure - E
Player Position: P1 at (1, 1)
Extra tile: I1
\end{PromptCode}
\end{tcolorbox}

\clearpage
\begin{tcolorbox}[recapprompt,title={Prompt for \Mbloxorz}] 
\raggedright
\setlength{\parindent}{0pt}
\setlength{\parskip}{0.3em}

You are solving a Bloxorz puzzle. A 1x1x2 rectangular block sits on a grid. Your goal is to roll the block so it stands UPRIGHT on the GOAL cell.

\textbf{RULES:}

- The block can be rolled UP, DOWN, LEFT, or RIGHT.\\
- The block has three orientations:\\
\quad * STANDING: upright on a single cell (1x1 footprint)\\
\quad * LYING HORIZONTALLY: flat, occupying two cells side by side (same row, adjacent columns)\\
\quad * LYING VERTICALLY: flat, occupying two cells stacked (same column, adjacent rows)\\
- When STANDING and you roll LEFT/RIGHT, the block tips over horizontally (occupies 2 cells in a row).\\
- When STANDING and you roll UP/DOWN, the block tips over vertically (occupies 2 cells in a column).\\
- When LYING HORIZONTALLY and you roll LEFT/RIGHT, the block stands up on one cell.\\
- When LYING HORIZONTALLY and you roll UP/DOWN, it slides (stays horizontal).\\
- When LYING VERTICALLY and you roll UP/DOWN, the block stands up on one cell.\\
- When LYING VERTICALLY and you roll LEFT/RIGHT, it slides (stays vertical).\\
- ALL cells occupied by the block must be on NORMAL, GOAL, or CRUMBLE tiles. If any cell is VOID or out of bounds, the block falls and you lose.\\
- You WIN when the block is STANDING on the GOAL cell.

\textbf{CRUMBLE TILE RULES:}

- CRUMBLE tiles (C) are single-use support tiles.\\
- A CRUMBLE tile is safe while the block is currently occupying it.\\
- Once the block fully leaves a CRUMBLE tile, that tile collapses permanently into a hole.\\
- A collapsed CRUMBLE tile behaves like VOID for the rest of the puzzle.\\
- You must plan around irreversible changes: crossing a CRUMBLE tile can remove that route for future backtracking.

\textbf{ELEVATOR RULES:}

- The board has TWO height levels: LOW (0) and HIGH (1). Some tiles are ELEVATED (at height 1).\\
- The block can only move between tiles at the SAME height level. Elevated tiles are like cliffs -- you cannot roll onto them from ground level or vice versa.\\
- ELEVATOR tiles (E) are special: they are accessible from BOTH height levels. When the block is STANDING (upright) on an elevator tile, it automatically toggles the block's height (\verb|LOW->HIGH| or \verb|HIGH->LOW|). The block must be STANDING -- lying on an elevator does NOT activate it.\\
- After an elevator toggles your height, you can now access tiles at the new height level.\\
- To WIN, the block must be STANDING on the GOAL at the SAME height as the goal tile.\\
- The board state is shown as TWO grids: one for LOW level and one for HIGH level. Tiles at the other height appear as '\#' (inaccessible). Elevator tiles appear on BOTH grids.\\
- Strategy: plan which elevators to use to reach the goal at the correct height.

\textbf{GRID SYMBOLS:}

\begin{PromptCode}
  . = Normal tile (safe)
  # = Void (hole - block falls if it touches this)
  G = Goal tile (stand upright here to win)
  C = Crumble tile (safe while occupied, collapses permanently after you leave it)
  E = Elevator tile (toggles block between LOW and HIGH height levels)
  B = Block (standing)
  H = Block (lying horizontally, both cells)
  V = Block (lying vertically, both cells)
\end{PromptCode}

Coordinates: (x, y) where x is column index (\verb|left->right|), y is row index (\verb|top->bottom|).

Directions: UP = toward y=0 (top of grid), DOWN = toward last y (bottom), LEFT = toward x=0, RIGHT = toward last x.

\textbf{MOVEMENT EXAMPLES (showing cell positions after each move):}

\begin{PromptCode}
  STANDING at (3,2):
    RIGHT -> LYING_H occupying (4,2) and (5,2)
    LEFT  -> LYING_H occupying (1,2) and (2,2)
    DOWN  -> LYING_V occupying (3,3) and (3,4)
    UP    -> LYING_V occupying (3,0) and (3,1)
  LYING_H at (4,2)-(5,2):
    RIGHT -> STANDING at (6,2)
    LEFT  -> STANDING at (3,2)
    DOWN  -> LYING_H at (4,3)-(5,3)  (slides, stays horizontal)
    UP    -> LYING_H at (4,1)-(5,1)  (slides, stays horizontal)
  LYING_V at (3,3)-(3,4):
    DOWN  -> STANDING at (3,5)
    UP    -> STANDING at (3,2)
    RIGHT -> LYING_V at (4,3)-(4,4)  (slides, stays vertical)
    LEFT  -> LYING_V at (2,3)-(2,4)  (slides, stays vertical)
\end{PromptCode}

\textbf{TASK}

Solve this Bloxorz puzzle in at most 16 moves.\\
You will receive the updated board state after each move.\\
Choose exactly one next move for this turn.

\textbf{OUTPUT}

Think step by step about the block's orientation after each move.\\
Your response should end with:

\begin{PromptCode}
```
<UP|DOWN|LEFT|RIGHT>
```
\end{PromptCode}

\textbf{Current board state:}

\begin{PromptCode}
--- LOW LEVEL (height 0) ---
{array}

--- HIGH LEVEL (height 1) ---
{array}

Block position: anchor at (x=6, y=2)
Block orientation: STANDING (upright on one cell)
Block occupies cells: (x=6, y=2)
Goal: (x=1, y=1)
Grid size: 7 cols (x) x 5 rows (y)
Block height level: LOW (0)
Elevator tiles (1): (x=4,y=0)
  Elevators toggle the block between LOW and HIGH when the block lands on them.
Elevated tiles (20): {coordinates}
Goal height level: HIGH (1)
Crumble tiles (5): (x=1,y=0), (x=5,y=0), (x=5,y=1), (x=4,y=4), (x=5,y=4)
  WARNING: CRUMBLE tiles are safe while occupied, but collapse into holes after the block fully leaves them.
\end{PromptCode}

(See the attached image for a visual representation of the board.)

\end{tcolorbox}

\newpage
\section{Experiment Details}
\label{sup:experiment_details}
\subsection{Experiment Configuration}

Table~\ref{tab:models} lists all models evaluated in this paper along with inference hyperparameters. Open-weight models are served locally via vLLM~\cite{kwon2023efficient} or SGLang~\cite{zheng2024sglang} on 16 Nvidia GH200 GPUs. Closed-weight models are accessed through their respective APIs. Unless otherwise noted, the maximum output length is set to 32{,}768 tokens.

\begin{table}[h]
\centering
\renewcommand{\arraystretch}{1.1}
\caption{MLLMs and corresponding hyperparameters. ``Reasoning effort'' controls the budget of reasoning tokens before the final response. \textsuperscript{*}For reasoning models, max tokens denotes the sum of reasoning and response tokens.}
\vspace{0.6em}
\small
\begin{tabular}{llcccc}
\toprule
\textbf{Model} & \textbf{Size} & \textbf{Release Date} & \textbf{Temp.} & \textbf{Reasoning Effort} & \textbf{Max Tokens\textsuperscript{*}} \\
\midrule
\textit{Open-weight models} \\
GLM-4.6V        & 106B  & Dec. 2025 & 0.8 & --     & 32{,}768 \\
Qwen3.6-27B     & 27B  & Apr. 2026 & 1.0 & --     & 32{,}768 \\
Llama-4-Scout   & 109B & Apr. 2025 & 0.6 & --     & 32{,}768 \\
Qwen3-VL-235B   & 235B & Sep. 2025 & 0.8 & --     & 32{,}768 \\
Qwen3.5-397B    & 397B & Feb. 2026 & 0.6 & --     & 32{,}768 \\
Gemma-4-31B     & 31B  & Apr. 2026 & 1.0 & --     & 32{,}768 \\
Kimi-K2.6       & 1.1T & Apr. 2026 & 1.0 & --     & 32{,}768 \\
\midrule
\multicolumn{5}{@{}l}{\textit{Closed-weight models (API)}}\\
Grok-4.3        & --   & May 2026  & -- & --     & -- \\
Gemini-3.1-pro  & --   & Feb. 2026 & -- & medium & -- \\
GPT-5.4         & --   & Mar. 2026 & -- & medium & -- \\
\bottomrule
\end{tabular}
\label{tab:models}
\end{table}

\subsection{Human Study}
\label{sup:human_study}

We conduct a human performance study to establish reference baselines for all three tasks in \DatasetName{}. The study is administered through a custom web-based interface built with Gradio, accessible to participants via a shared URL. Participants provide informed consent before beginning and may withdraw at any time.

\paragraph{Participants.}
We recruit $8$ experienced players familiar with spatial puzzle games. Each participant completes tasks across all three environments (\Mcube{}, \Mmaze{}, and \Mbloxorz{}), with each environment containing both a perception sub-task and a planning sub-task.

\paragraph{Protocol.}
For each environment, participants first complete practice trials with immediate correctness feedback, followed by scored trials without feedback. The detailed trial structure is as follows:

\begin{itemize}[nosep,leftmargin=*]
    \item \textbf{\Mcube{} Perception}: practice + scored trials. The participant extracts a $5 \times 5$ binary grid from an isometric 3D rendering of a single puzzle piece.
    \item \textbf{\Mcube{} Planning}: practice + easy (2 missing pieces) + hard (4 missing pieces) trials. The participant places missing pieces on the cube, specifying the face assignment and orientation for each.
    \item \textbf{\Mmaze{} Perception}: practice + scored trials. The participant identifies the tile type (I/L/T) and rotation (0--3) of a highlighted cell on the labyrinth board.
    \item \textbf{\Mmaze{} Planning}: practice + easy (depth 2) + hard (depth 4) trials. The participant inserts tiles and moves the pawn to reach the target, with a step budget of 6 (easy) or 12 (hard) individual actions.
    \item \textbf{\Mbloxorz{} Perception}: practice + scored trials. The participant identifies the block position, orientation, and goal location from an isometric board rendering.
    \item \textbf{\Mbloxorz{} Planning}: practice + easy + hard trials. The participant rolls the block to stand upright on the goal cell, with a budget of 16 (easy) or 24 (hard) moves.
\end{itemize}

\paragraph{Interface.}
The Gradio-based interface presents each trial as an interactive form: participants view the rendered game state and submit their responses through task-specific input widgets (dropdown menus for tile type and rotation, grid editors for binary grids, directional buttons for block movement, etc.). Practice trials display immediate feedback (correct/incorrect with the ground-truth answer), while scored trials simply advance to the next problem. A progress tracker shows completion status for each sub-task.

\begin{figure}[h]
    \centering
    \includegraphics[width=\linewidth]{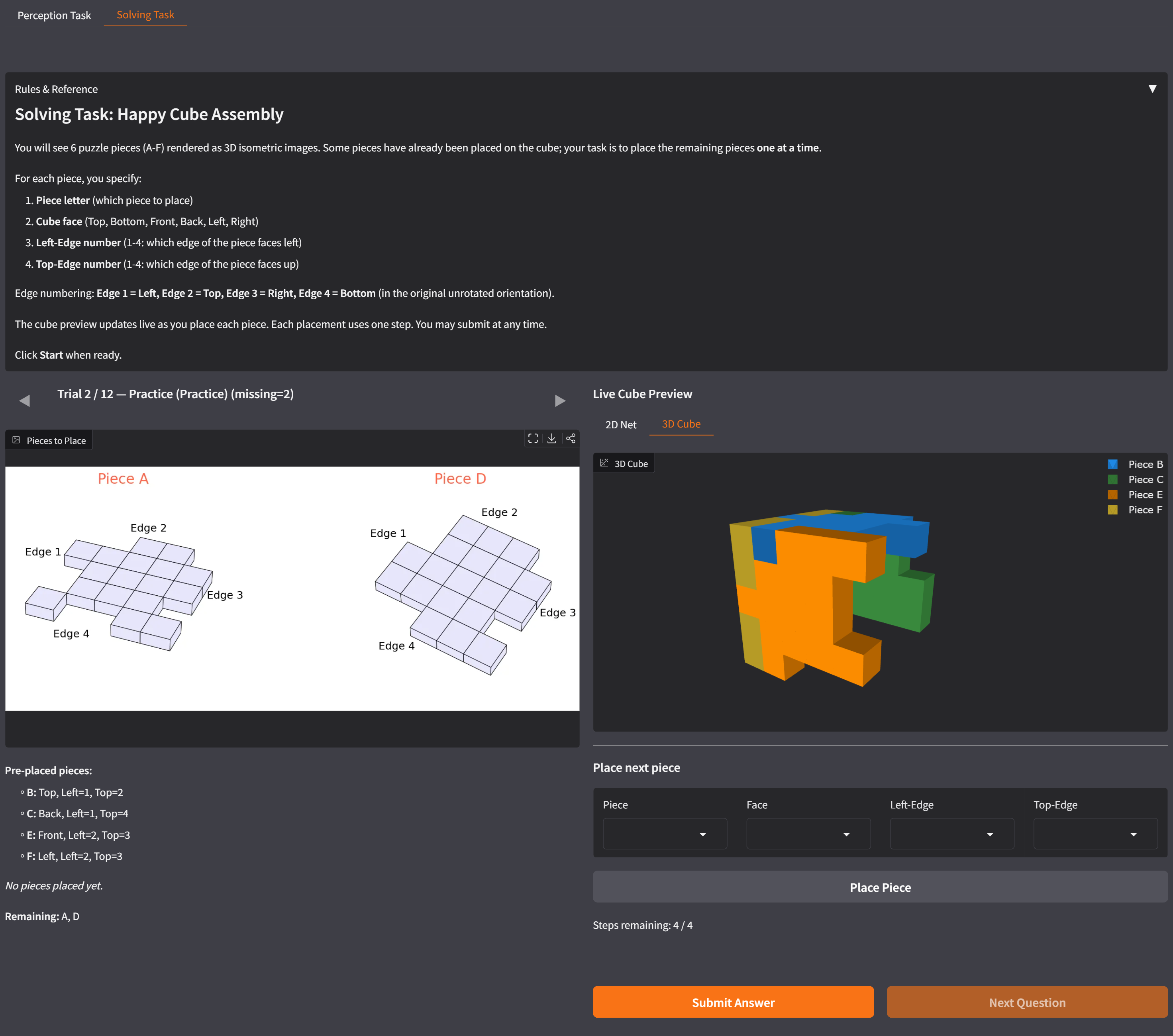}
    \caption{\textbf{Human Study Interface.} Screenshot of the \Mcube{} planning task in the Gradio-based web application. Participants inspect the puzzle pieces and the partially assembled cube, then select a piece, target face, and orientation to complete the assembly.}
    \label{fig:human_study_interface}
\end{figure}

\paragraph{Human performance.}
We evaluated eight participants per task.
Table~\ref{tab:human_performance} reports the mean success rate and standard deviation across participants. Hard instances exhibit larger variability across the participants, particularly in \Mcube{} and \Mbloxorz{}. Nevertheless, the decline in human performance from easy to hard remains substantially smaller than that observed for frontier models.

\begin{table}[h]
    \centering
    \caption{Human performance on \DatasetName{} with eight participants
    per task, reported as mean success rate $\pm$ standard deviation (\%).}
    \label{tab:human_performance}
    \vspace{1em}
    \begin{tabular}{lcc}
        \toprule
        Environment & Easy & Hard \\
        \midrule
        \Mcube   & $86.6 \pm 10.5$ & $85.0 \pm 23.3$ \\
        \Mmaze   & $82.5 \pm 20.1$ & $78.8 \pm 25.3$ \\
        \Mbloxorz & $78.6 \pm 22.9$ & $62.5 \pm 31.1$ \\
        \bottomrule
    \end{tabular}
\end{table}

\subsection{Spatial State Parsing Tasks}
\label{sup:parsing_tasks}

\begin{figure}[h]
    \centering
    \includegraphics[width=\linewidth]{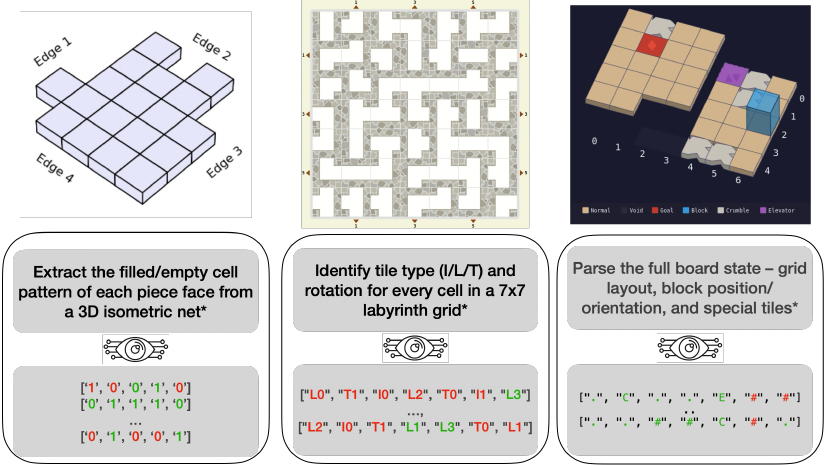}
    \caption{\textbf{Spatial State Parsing Tasks.} Example inputs and expected structured outputs for each parsing task: \Mcube{} (binary grid extraction), \Mmaze{} (tile type and rotation), and \Mbloxorz{} (grid layout, block position, and orientation).}
    \label{fig:parsing_tasks_appendix}
\end{figure}

Section~\ref{subsec:visual_abstract} introduces three atomic spatial parsing tasks that isolate visual-to-spatial abstraction from downstream planning. Here we describe each task in detail and illustrate the expected output format (Figure~\ref{fig:parsing_tasks_appendix}).

\paragraph{\Mcube{} Parsing.}
Given a 3D isometric rendering of a single puzzle piece, the model must extract its $5 \times 5$ binary grid indicating which cells are filled (\texttt{1}) and which are empty (\texttt{0}). The central $3 \times 3$ region is always filled; the difficulty lies in correctly identifying the bump-and-gap pattern along the four boundary edges from the isometric viewpoint. Per-cell accuracy is computed over all 25 cells.

\paragraph{\Mmaze{} Board Parsing.}
Given a flat 2D rendering of the $7 \times 7$ labyrinth board, the model must output a structured grid where each cell specifies the tile type (I, L, or T) and its rotation ($0$--$3$). Each tile type is defined by its corridor openings: I-tiles have two opposing corridors, L-tiles have two adjacent corridors forming a corner, and T-tiles have three corridors. The rotation index encodes the absolute orientation of the corridor pattern. Per-cell accuracy requires both tile type and rotation to be correct.

\paragraph{\Mbloxorz{} Parsing.}
Given a 3D isometric rendering of a \Mbloxorz{} board, the model must extract three components: (1) the grid layout, a binary matrix indicating which cells belong to the platform (\texttt{.}) versus void (\texttt{\#}), with special markers for elevators and goal cells; (2) the block position, reporting which cell(s) the block currently occupies; and (3) the block orientation (upright, horizontal, or vertical). Per-cell accuracy is computed over the grid layout; block position and orientation are evaluated as exact match.

\clearpage
\section{Additional Results}
\label{sup:additional_results}

\textbf{Output Token Usage.}
Table~\ref{tab:output_token_usage} reports the average output tokens per turn during evaluation. The results show that models spend substantial reasoning effort on each single step, with Kimi-K2.6 averaging 32.5k output tokens  per turn on easy \Mcube{} instances. 
Interestingly, for \Mcube and \Mmaze, several models generate fewer output tokens in the hard setting than in the easy setting, despite the longer planning horizons.
We hypothesize that this reduction stems from difficulties in visual-to-spatial abstraction. Without symbolic state descriptions, models may struggle to construct reliable representations of the current state, limiting their ability to reason through subsequent state transitions and develop detailed plans. This observation is consistent with our diagnostic findings that visual abstraction constitutes a major bottleneck in solving \DatasetName{} tasks.

\begin{table}[h]
\centering
\caption{Average output tokens per turn in the multi-turn evaluation.}
\vspace{0.5em}
\label{tab:output_token_usage}
\small
\setlength{\tabcolsep}{6pt}
\renewcommand{\arraystretch}{1.08}
\begin{tabular}{lrrrrrr}
\toprule
& \multicolumn{2}{c}{CUBE} & \multicolumn{2}{c}{MAZE} & \multicolumn{2}{c}{BLOX} \\
\cmidrule(lr){2-3}\cmidrule(lr){4-5}\cmidrule(lr){6-7}
Model & Easy & Hard & Easy & Hard & Easy & Hard \\
\midrule
GLM-4.6V & 8,633 & 2,304 & 3,716 & 1,440 & 1,942 & 1,682 \\
Gemma-4-31B & 11,690 & 9,360 & 8,321 & 5,948 & 3,140 & 4,506 \\
Qwen3.6-27B & 19,622 & 11,448 & 14,607 & 8,171 & 8,122 & 7,879 \\
Qwen3-VL-235B & 16,084 & 7,688 & 7,968 & 3,476 & 3,033 & 1,968 \\
Qwen3.5-397B & 29,750 & 15,427 & 21,057 & 15,026 & 6,712 & 8,087 \\
Kimi-K2.6 & 32,538 & 20,422 & 25,818 & 14,311 & 5,371 & 17,398 \\
\midrule
Grok-4.3 & 9,383 & 1,982 & 7,577 & 1,739 & 1,534 & 2,599 \\
Gemini-3.1-Pro & 9,457 & 14,461 & 5,153 & 9,251 & 1,383 & 6,954 \\
GPT-5.4 & 13,664 & 5,238 & 6,856 & 5,063 & 1,529 & 5,024 \\
\bottomrule
\end{tabular}
\end{table}

\begin{figure}[h]
    \centering
    \includegraphics[width=0.5\linewidth]{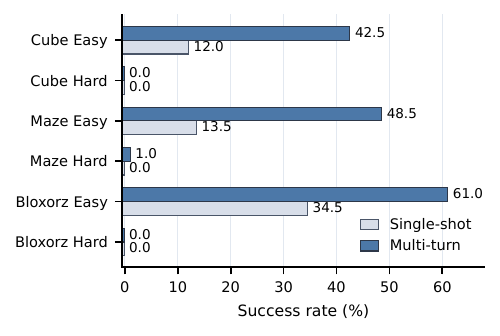}
    \caption{Single-turn versus multi-turn success rate for Gemma-4-31B across \DatasetName{} environments.}
    \label{fig:single_turn_vs_multi_turn}
\end{figure}

\textbf{Does Interaction Matter?}
Each \DatasetName{} task can also be evaluated in a single-turn setting, where the agent must generate a complete action plan from the initial observation alone, without intermediate environment feedback. Figure~\ref{fig:single_turn_vs_multi_turn} compares single-turn and multi-turn performance for Gemma-4-31B. 
Multi-turn interaction substantially improves performance in the easy settings, increasing success from 12.0\% to 42.5\% on \Mcube, from 13.5\% to 48.5\% on \Mmaze, and from 34.5\% to 61.0\% on \Mbloxorz.
The difference becomes less prominent in the hard settings, where performance remains at 0.0\% on \Mcube{} and \Mbloxorz{} and improves only marginally from 0.0\% to 1.0\% on \Mmaze{}. 
These results suggest that interaction helps models recover from imperfect intermediate decisions, but is still insufficient for solving harder visually grounded spatial reasoning tasks in \DatasetName.

\newpage
\clearpage
\section*{NeurIPS Paper Checklist}

\begin{enumerate}

\item {\bf Claims}
    \item[] Question: Do the main claims made in the abstract and introduction accurately reflect the paper's contributions and scope?
    \item[] Answer: \answerYes{} 
    \item[] Justification: The abstract and Introduction describe \DatasetName as a diagnostic benchmark for visual-to-spatial abstraction and long-horizon spatial planning, and the reported claims are supported by the main results and diagnostic ablations in Sections~\ref{sec:experiment}.


\item {\bf Limitations}
    \item[] Question: Does the paper discuss the limitations of the work performed by the authors?
    \item[] Answer: \answerYes{} 
    \item[] Justification: We discuss the limitation of the paper in Section~\ref{sec:conclusion}.

\item {\bf Theory assumptions and proofs}
    \item[] Question: For each theoretical result, does the paper provide the full set of assumptions and a complete (and correct) proof?
    \item[] Answer:\answerNA{}. 
    \item[] Justification: The paper does not present theoretical results, theorems, or formal proofs.

    \item {\bf Experimental result reproducibility}
    \item[] Question: Does the paper fully disclose all the information needed to reproduce the main experimental results of the paper to the extent that it affects the main claims and/or conclusions of the paper (regardless of whether the code and data are provided or not)?
    \item[] Answer: \answerYes{} 
    \item[] Justification: The paper describes the benchmark design, evaluation details, and metrics. Code and data are also available for reproducing the results.    

\item {\bf Open access to data and code}
    \item[] Question: Does the paper provide open access to the data and code, with sufficient instructions to faithfully reproduce the main experimental results, as described in supplemental material?
    \item[] Answer: \answerYes{} 
    \item[] Justification: Code and data are provided with public access.

\item {\bf Experimental setting/details}
    \item[] Question: Does the paper specify all the training and test details (e.g., data splits, hyperparameters, how they were chosen, type of optimizer) necessary to understand the results?
    \item[] Answer: \answerYes{}{} 
    \item[] Justification: The paper explains the experiment details in Section~\ref{sec:experiment} and Appendix~\ref{sup:experiment_details}.

\item {\bf Experiment statistical significance}
    \item[] Question: Does the paper report error bars suitably and correctly defined or other appropriate information about the statistical significance of the experiments?
    \item[] Answer: \answerNo{} 
    \item[] Justification: We report success rates over fixed benchmark splits but do not report repeated-run error bars or statistical significance tests. Running multiple independent repetitions for all frontier MLLMs is prohibitively expensive due to API and compute budget constraints; instead, we focus on controlled comparisons across observation format and planning horizon, where the main conclusions are supported by large performance gaps.

\item {\bf Experiments compute resources}
    \item[] Question: For each experiment, does the paper provide sufficient information on the computer resources (type of compute workers, memory, time of execution) needed to reproduce the experiments?
    \item[] Answer: \answerYes{} 
    \item[] Justification: The compute resources are discussed in Section~\ref{sec:experiment} and Appendix~\ref{sup:experiment_details}.
    
\item {\bf Code of ethics}
    \item[] Question: Does the research conducted in the paper conform, in every respect, with the NeurIPS Code of Ethics \url{https://neurips.cc/public/EthicsGuidelines}?
    \item[] Answer:  \answerYes{}. 
    \item[] Justification: The work uses synthetic puzzle environments and existing MLLMs for benchmark evaluation, and does not rely on private, sensitive, or scraped personal data. Human evaluation details are handled according to the applicable institutional requirements.

\item {\bf Broader impacts}
    \item[] Question: Does the paper discuss both potential positive societal impacts and negative societal impacts of the work performed?
    \item[] Answer: \answerYes{} 
    \item[] Justification: We've discussed the broader impact of the paper.
    
\item {\bf Safeguards}
    \item[] Question: Does the paper describe safeguards that have been put in place for responsible release of data or models that have a high risk for misuse (e.g., pre-trained language models, image generators, or scraped datasets)?
    \item[] Answer: \answerNA{} 
    \item[] Justification: The paper does not release high-risk generative models, scraped datasets, private data, or assets with an obvious dual-use misuse pathway; the benchmark instances are synthetic spatial puzzle environments.

\item {\bf Licenses for existing assets}
    \item[] Question: Are the creators or original owners of assets (e.g., code, data, models), used in the paper, properly credited and are the license and terms of use explicitly mentioned and properly respected?
    \item[] Answer: \answerYes{} 
    \item[] Justification: We cite all evaluated models through citations in the manuscript, and our use of closed-weight models follows their public API access terms while open-weight models are used under their released licenses. \DatasetName instances are synthetically generated by our environment code and do not redistribute third-party datasets.

\item {\bf New assets}
    \item[] Question: Are new assets introduced in the paper well documented and is the documentation provided alongside the assets?
    \item[] Answer: \answerYes{}{} 
    \item[] Justification: The paper introduces \DatasetName as a new benchmark and provides the data and code with detailed explanation and instruction.

\item {\bf Crowdsourcing and research with human subjects}
    \item[] Question: For crowdsourcing experiments and research with human subjects, does the paper include the full text of instructions given to participants and screenshots, if applicable, as well as details about compensation (if any)? 
    \item[] Answer: \answerYes{} 
    \item[] Justification: The details of the human study are provided in Appendix~\ref{sup:experiment_details}.

\item {\bf Institutional review board (IRB) approvals or equivalent for research with human subjects}
    \item[] Question: Does the paper describe potential risks incurred by study participants, whether such risks were disclosed to the subjects, and whether Institutional Review Board (IRB) approvals (or an equivalent approval/review based on the requirements of your country or institution) were obtained?
    \item[] Answer: \answerYes{} 
    \item[] Justification: The human evaluation was conducted with approval from the relevant parties involved in the study. Participants were informed that they would solve synthetic spatial puzzle tasks for research evaluation, participation was voluntary, and only aggregate, anonymized performance statistics are reported.

\item {\bf Declaration of LLM usage}
    \item[] Question: Does the paper describe the usage of LLMs if it is an important, original, or non-standard component of the core methods in this research? Note that if the LLM is used only for writing, editing, or formatting purposes and does \emph{not} impact the core methodology, scientific rigor, or originality of the research, declaration is not required.
    \item[] Answer: \answerYes{} 
    \item[] Justification: MLLMs are central experimental subjects in this work, and the paper describes the evaluated open-weight and closed-weight models in Sections~\ref{sec:experiment}.

\end{enumerate}


\end{document}